\documentclass{agujournal2019}
\usepackage[T2A]{fontenc}
\usepackage{fix-cm}
\usepackage{url}
\usepackage{soul}
\usepackage{booktabs}

\usepackage{amsfonts}
\usepackage{xfrac}
\usepackage{wrapfig}
\usepackage{multirow}
\usepackage{changepage} 

\definecolor{green2}{rgb}{0.0, 0.70, 0.0}

\draftfalse

\journalname{Journal of Advances in Modeling Earth Systems (JAMES)}

\usepackage{cleveref}

\begin{document}

\title{Local Off-Grid Weather Forecasting with Multi-Modal Earth Observation Data}

\authors{Qidong Yang\affil{1}, Jonathan Giezendanner\affil{1},
Daniel Salles Civitarese\affil{2}, Johannes Jakubik\affil{2},
Eric Schmitt\affil{3}, Anirban Chandra\affil{3}, Jeremy Vila\affil{3}, Detlef Hohl\affil{3}, Chris Hill\affil{1}, Campbell Watson\affil{2}, and Sherrie Wang\affil{1}}

\affiliation{1}{Massachusetts Institute of Technology}
\affiliation{2}{IBM Research}
\affiliation{3}{Shell Information Technology International Inc.}

\correspondingauthor{Qidong Yang, Jonathan Giezendanner, Sherrie Wang}{\{qidong, jgiezend, sherwang\}@mit.edu}

\begin{keypoints}
\item There is a systematic bias between gridded numerical weather products and off-grid local weather station measurements.
\item A model combining gridded numerical weather products and off-grid weather measurements can make accurate and localized weather forecasts for a given lead time.
\item Transformer is one of the best performing models for off-grid local weather forecasting due to its flexible attention mechanism.
\end{keypoints}

\begin{abstract}
      Urgent applications like wildfire management and renewable energy generation require precise, 
localized weather forecasts near the Earth's surface.
However, forecasts produced by machine learning models or numerical weather prediction systems are typically generated on large-scale regular grids, where direct downscaling fails to capture fine-grained, near-surface weather patterns.
In this work, we propose a multi-modal transformer model trained end-to-end to downscale gridded forecasts to off-grid locations of interest.
Our models directly combine local historical weather observations (e.g., wind, temperature, dewpoint) with gridded forecasts to produce locally accurate predictions at various lead times.
Multiple data modalities are collected and concatenated at station-level locations, treated as a token at each station.
Using self-attention, the token corresponding to the target location aggregates information from its neighboring tokens.
Experiments using weather stations across the Northeastern United States show that our model outperforms a range of data-driven and non-data-driven off-grid forecasting methods.
They also reveal that direct input of station data provides a phase shift in local weather forecasting accuracy, reducing the prediction error by up to 80\% compared to pure gridded data based models.
This approach demonstrates how to bridge the gap between large-scale weather models and locally accurate forecasts to support high-stakes, location-sensitive decision-making. \end{abstract}

\section*{Plain Language Summary}
In our research, we developed a new AI approach to make weather forecasts more accurate for specific locations, which is vital for urgent needs like wildfire management and renewable energy production.
Traditional weather forecasts work on large grids covering wide areas. This makes them less reliable for specific spots on the ground where precise predictions matter most.
Our method uses a special type of AI (a multi-modal transformer) that learns from both traditional forecasts and actual weather measurements from local weather stations. By combining these different data sources, our AI can "correct" the large-scale forecasts for specific locations.
For example, our system looks at historical temperature, wind, and moisture readings from a network of weather stations, then uses this local knowledge to refine predictions for exact spots people care about. 
Our AI also skillfully considers how nearby weather stations relate to each other, giving more weight to stations that provide relevant information.
When we tested our approach across the Northeastern United States, it provided more accurate local forecasts than existing methods. 
This improvement could help decision-makers in high-stakes situations like planning for wildfire response or managing renewable energy systems, where knowing precise local weather conditions can make a critical difference.

\section{Introduction}
In recent years, machine learning (ML) has been widely used in weather forecasting applications. 
This popularity stems from its fast inference speed and ability to model complex physical dynamics directly from data. 
Some high-profile ML weather forecasting models include FourCastNet~\cite{pathak2022}, GraphCast~\cite{lam2023}, and Pangu-Weather~\cite{bi2023}. 
These ML weather models can generate forecasts thousands of times faster than traditional numerical weather prediction (NWP) models, 
while at the same time being more accurate and flexible, freed from the NWP model's sometimes restrictive physical constraints~\cite{pathak2022,lam2023,kochkov2024}.

\begin{figure}[ht]
    \begin{center}
    \includegraphics[width=0.8\linewidth]{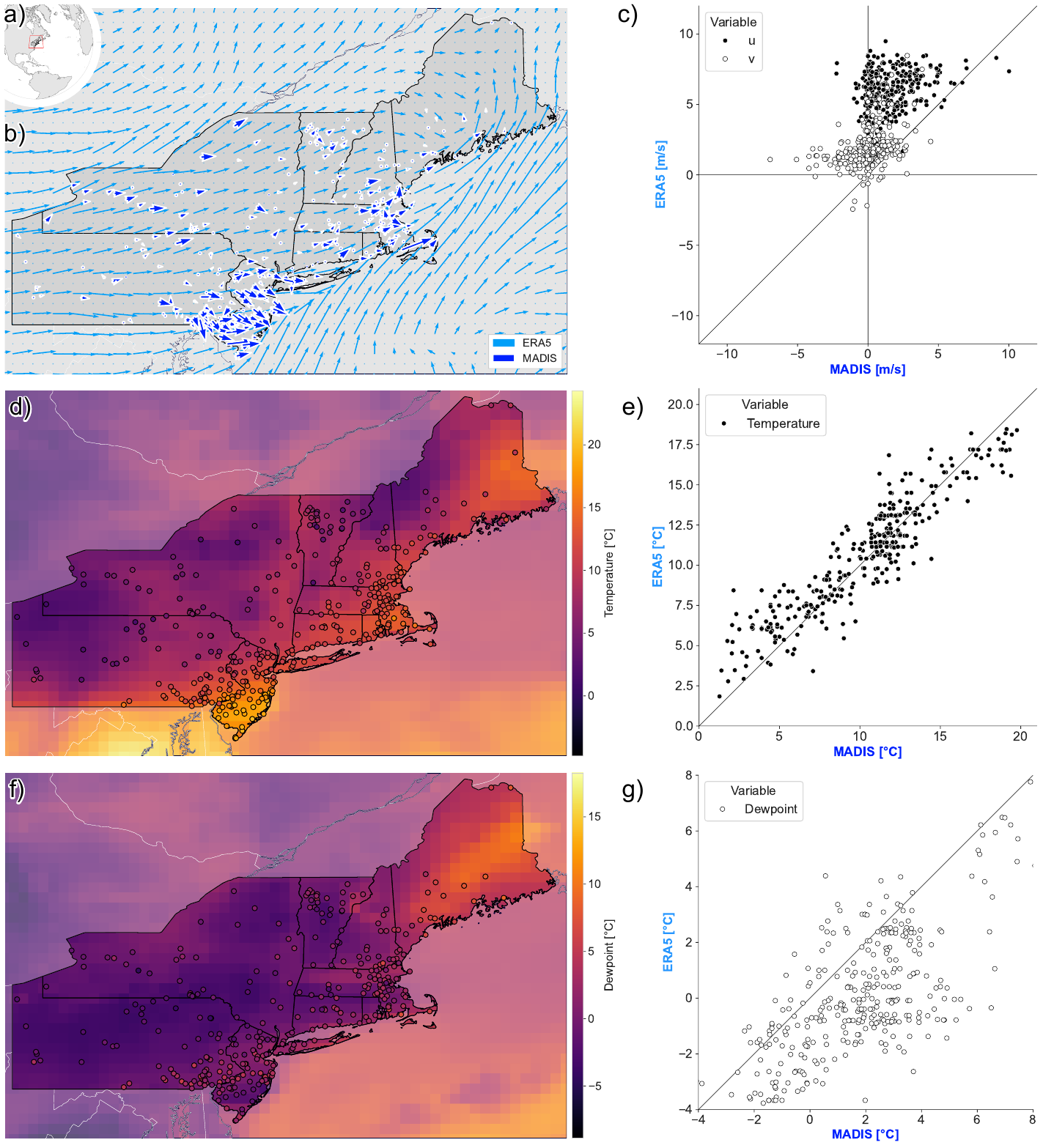}
    \end{center}
    \caption{\textbf{Gridded reanalysis data like ERA5 do not capture localized, near-surface weather dynamics.} 
    (a) Our study area is the Northeastern United States. 
    (b) Wind field for April 18$^{\text{th}}$, 2023 18:00--19:00, with ERA5 global reanalysis data in light blue and wind measured by weather stations (MADIS) in dark blue. 
    (c) Scatterplot of MADIS vs. ERA5 for the same time, separated into $u$ ($\rho^2 = 0.07$) and $v$ ($\rho^2 = 0.12$) components of wind.
    The ERA5 data is linearly interpolated to the locations of the MADIS weather stations.
    (d) and (e) are the same plots for temperature ($\rho^2 = 0.85$). (f) and (g) show dewpoint ($\rho^2 = 0.46$).
    }
    \label{fig:methods:data:era5vsmadis}
\end{figure}

To date, most ML weather models have been trained with gridded numerical weather reanalysis products like ERA5~\cite{hersbach2020}. 
However, reanalysis products have been shown to have a systematic bias relative to the weather station measurements~\cite{ramavajjala2023}. 
We verify the existence of a substantial bias in ERA5's near-surface wind estimates in Figure~\ref{fig:methods:data:era5vsmadis} (bias in single-time snapshot) and Figure~\ref{fig:methods:data:era5vsmadis_distribution} (bias in one-year statistics). 
ERA5 systematically overestimates the inland near-surface wind speed and is much smoother across space than the actual wind field as measured by weather stations. 
ML models trained to predict numerical weather reanalysis inherit this significant bias and are unable to make accurate localized predictions. 

This presents a challenge, as accurate off-grid weather forecasts are critical for applications like wildfire management and sustainable energy generation. 
To bridge this gap, we explore ML models to forecast localized weather patterns.
These models digest multi-modal Earth observation data, including gridded numerical weather forecast products and local weather station measurements, 
to make accurate forecasts at irregularly spaced off-grid points.

First, we curate a dataset that contains both numerical weather products (ERA5 and HRRR) and local weather station observations (MADIS), 
spanning 2019--2023 and covering the Northeastern United States.
On top of the dataset, a transformer model is trained to make forecasts at each weather station.
It tokenizes the weather measurements at each station and concatenates them with numerical weather estimates at nearby grid points.
It processes the multi-modal input tokens with a stack of transformer encoder blocks,
then makes a final prediction at each station by decoding the output of the last encoder block with a prediction head.
As a competitive baseline, we also explore a graph neural network (GNN) to forecast at each weather station, as
GNNs are widely used in multi-modal off-grid settings~\cite{brandstetter2023} and in weather forecasting~\cite{lam2023}.
Here, the GNN is built on a heterogeneous graph, where NWP grid points and off-grid weather stations are treated as two different types of nodes.
The GNN operates on this graph and makes forecasts at each weather station. 
When making predictions at a station location, the GNN aggregates information from neighboring weather stations and numerical weather grid points using message passing~\cite{gilmer2017}.

These two methods preserve the irregular geometry of the off-grid stations and theoretically infinite spatial resolution,
and ingest both gridded NWP and local weather station measurements.
As a result, the prediction is informed by both large-scale atmospheric dynamics and local weather patterns. 
We evaluate our models' ability to forecast real data from weather stations, focusing on temperature, dewpoint, and wind.
Our transformer model outperforms a variety of other off-grid forecasting methods, including ERA5 interpolation and time series forecasting without spatial context.
It also outperforms the GNN model by a significant margin, indicating that the transformer model is able to better capture the complex spatial relationships between weather stations.

Our contributions can be summarized as the following:
\begin{enumerate}
    \item We compile and release a multi-modal weather dataset incorporating both off-grid MADIS weather stations and gridded ERA5 and HRRR. 
    The dataset covers the Northeastern US from 2019--2023 and includes a comprehensive list of weather variables. 
    \item We verify, using our dataset, the systematic bias between gridded numerical weather products (ERA5 and HRRR) and off-grid local weather station measurements (MADIS).
    \item We propose a multi-modal transformer to model local weather dynamics at the station level, taking advantage of both gridded numerical weather products and weather station observations.
    One separate forecasting model is trained for each given lead time.
    \item We evaluate our models against a range of data-driven and non-data-driven off-grid weather forecasting methods. 
    Amongst those, our transformer model achieves the best performance. 
    It decreases the average error by 36\% compared to the best performing ML baseline model, which in turn reduces the mean forecast error by 34\% compared to the best performing non-ML baseline model (interpolated HRRR analysis).
    \item We conduct an ablation of inputs and observed that all ML models improve significantly when NWP are used as inputs.
    For example, a transformer with HRRR input achieves 50\% of the error of a transformer without HRRR, indicating that---even in the presence of historical station data---large-scale atmospheric dynamics inform local weather patterns.
\end{enumerate}

\section{Related Work}
\textbf{Dynamical Downscaling}
The mismatch between the large-scale weather/climate simulation and the local observations is a long-standing problem in atmospheric sciences.
Downscaling techniques have been widely used to alleviate the mismatch by improving the resolution and debiasing of large-scale weather/climate simulations.
One type of downscaling technique is dynamical downscaling, which uses high-resolution regional models or limited-area models that are nested within global numerical models to generate finer-scale simulations.
This approach explicitly solves physical equations governing atmospheric dynamics and thermodynamics at a higher resolution, 
allowing for a more detailed representation of local topography, land-atmosphere interactions, and mesoscale weather patterns. 
Prominent models used for dynamical downscaling include the Weather Research and Forecasting (WRF) model~\cite{skamarock} and the Regional Climate Model (RegCM)~\cite{giorgi2012}. 
Despite its ability to provide physically consistent and high-fidelity representations of local dynamics, dynamical downscaling is computationally expensive, requiring significant resources for running high-resolution simulations over extended periods. 
Moreover, biases and uncertainties from the driving global model can propagate into the regional model, affecting accuracy.

\textbf{Statistical Downscaling}
Statistical downscaling, on the other hand, leverages statistical or machine learning techniques to establish empirical relationships between large-scale coarse-resolution model outputs and local weather observations. 
Traditional statistical downscaling methods include linear regression models~\cite{wilby1998}, canonical correlation analysis~\cite{storch1993}, and analog method~\cite{zorita1999}. 
These methods are extensively evaluated and compared in \citeA{hernanz2022}. 
More recently, deep learning-based approaches, such as convolutional neural networks (CNNs)~\cite{vandal2017} and generative adversarial networks (GANs)~\cite{stengel2020}, have been explored for downscaling tasks. 
Compared to dynamical downscaling, statistical downscaling is computationally more efficient and can be easily adapted to different regions and variables. 
However, its performance depends on the availability of high-quality training data, and it may struggle with capturing physically consistent interactions beyond the training data.

\textbf{Gridded Weather Forecasting}
Weather forecasting has long been a challenging problem in atmospheric sciences, with efforts dating back centuries. 
Since the advent of numerical weather prediction (NWP) in the mid-20th century, most forecast simulations have been conducted on a regular grid, dividing the atmosphere into evenly spaced discrete points to solve complex partial differential equations. 
This grid-based approach has remained the foundation of many numerical weather forecasting models such as the Integrated Forecast System~\cite{ECMWF2022} and High-Resolution Rapid Refresh~\cite{dowell2022}. 
In recent years, machine learning (ML) has gained traction as a promising tool in weather forecasting~\cite{bauer2015}, offering new techniques to improve accuracy and computational efficiency.
These ML weather models can be roughly divided into two categories: specialized models and foundation models. 
FourCastNet~\cite{pathak2022}, GraphCast~\cite{lam2023}, Pangu-Weather~\cite{bi2023}, AIFS~\cite{lang2024} and NeuralGCM~\cite{kochkov2024} are specialized models which are trained directly to make weather forecasts. 
In contrast, AtmoRep~\cite{lessig2023}, ClimaX~\cite{nguyen2023}, Aurora~\cite{bodnar2024} and Prithvi WxC~\cite{schmude2024} are foundation models that are first trained with a self-supervision task and then fine-tuned for weather forecasting.
However, the training data for ML models largely stem from traditional gridded numerical simulations such as ERA5~\cite{hersbach2020}. 
As a result, the ML models themselves still typically maintain the grid-based paradigm even within their more modern forecasting approach. 
One major disadvantage of gridded weather forecasting is that it is usually limited by its fixed resolution such that it cannot accurately reflect fine-grained local weather patterns (although efforts towards limited area modeling and storm-scale emulation have recently been made~\cite{oskarsson2023,pathak2024,flora2025,yang2020,yang2022}). 
Other works focusing on increasing the forecast resolution~\cite{harder2023,yang2023,prasad2024} exist, but their methods are mostly tested on synthetic datasets.
Work meant to correct ERA5 forecasts exists~\cite{mouatadid2023}, but focus on sub-seasonal forecast at coarse spatial resolution rather than local weather forecasting.
In this work, we propose multi-modal deep learning models which can effectively downscale gridded weather forecasts to match real-world local weather dynamics.

\textbf{Off-Grid Weather Forecasting}
Even though gridded weather forecasting is the main focus of the ML community, there have been several attempts to forecast weather off-grid. 
\citeA{bentsen2023} applied a GNN to forecast wind speed at 14 irregularly spaced off-shore weather stations, each of which was treated as a node within the graph. 
The model input is the historical trajectory of weather variables recorded at each station. 
This work has two limitations: the forecasting region is small, only covering 14 stations, and it only considers a single input modality of station historical measurements. 
MetNet-3~\cite{andrychowicz2023} takes another approach to off-grid weather forecasting. 
It trains a U-Net-like transformer~\cite{ronneberger2015} model that takes multi-modal inputs including weather station observations, satellite imagery, and assimilation products to predict weather at stations. 
However, both input and output station data are re-gridded to a high resolution mesh (4 km $\times$ 4 km), which distorts the off-grid data's original granularity.
Recently, \citeA{allen2025} trained an end-to-end ML weather forecasting model 
that first fuses multi-modal earth observation data onto a regular grid;
then makes time stepping into the future in the gridded space producing ERA5 like forecasts;
lastly, downscales the gridded weather forecast to individual off-grid station locations.
In contrast, our method produces one-step off-grid forecasts at each given lead time without temporal rollout by directly downscaling ready numerical forecasts.

\textbf{Graph Neural Network for Physical Simulation}
Graph neural networks (GNNs) are a type of deep learning model designed to operate on data structured as graphs, 
where entities are represented as nodes and their relationships as edges. 
GNNs provide flexibility to process data with non-Euclidean structures. 
A GNN learns to capture relationships between nodes by iteratively passing and aggregating information between neighboring nodes, 
and updating node representations based on their connections. 
Recently, GNNs have been widely used in physical system simulation. 
For example, the 2D Burgers' equation can be effectively solved on both a regular and an irregular mesh with GNNs such as MAgNet~\cite{boussif2022} and MPNN~\cite{brandstetter2023}. 
\citeA{sanchez-gonzalez2020} used a GNN to simulate particle dynamics in a wide variety of physical domains, involving fluids, rigid solids, and deformable materials interacting with one another. 
GraphCast~\cite{lam2023} even showed that a 2D GNN is capable of simulating a global gridded atmospheric system. 
These successful use cases of GNNs motivate us to apply a graph network to our task for localized off-grid weather forecasting.

\textbf{Transformer as Building Blocks for Weather Models}
Transformers~\cite{vaswani2023a,dosovitskiy2021a} have become the fundamental components of modern weather foundation models due to their ability to learn complex spatial-temporal relationships from large-scale atmospheric data. 
These models, such as AtmoRep~\cite{lessig2023} and ClimaX~\cite{nguyen2023}, leverage transformers to pretrain on diverse meteorological datasets, enabling them to generalize across various forecasting tasks. 
One key advantage of transformers is their flexibility in handling irregularly spaced data, making them well-suited for off-grid applications. 
Specifically, transformers treat irregularly spaced data as a sequence of tokens, and use self-attention mechanisms to process them without relying on a grid-based structure. 
Transformers provide a powerful and flexible solution for off-grid weather forecasting.
 
\section{Methods}
\begin{figure}[htbp]
    \begin{center}
        \includegraphics[width=1\columnwidth]{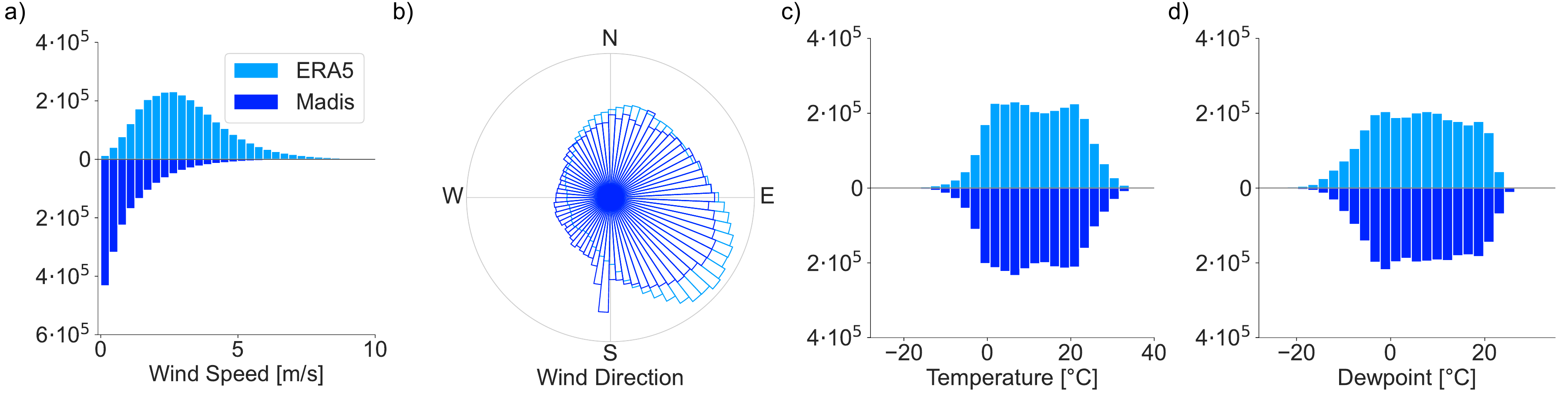}
    \end{center}
    \caption{\textbf{Comparison of data collected by weather stations (MADIS) and linearly interpolated global reanalysis data (ERA5).} For both ERA5 and MADIS data, (a) histogram of wind speed, (b) radial histogram of wind direction, (c) histogram of temperature and (d) histogram of dewpoint for the study region from January to December 2023. Large differences, especially in wind speed, are apparent between local wind observations and global wind products. ERA5, which is the target that most ML weather models emulate, does not capture local wind dynamics.}\label{fig:methods:data:era5vsmadis_distribution}
\end{figure}

In this work, we investigate two deep learning models for localized weather forecasting: a message passing neural network (MPNN,~\citeA{gilmer2017, pfaff2021}) and a transformer ~\cite{vaswani2023a,dosovitskiy2021a}.
Both models ingest multi-modal Earth observation data and are trained to forecast weather at the station level with the aid of large-scale weather predictions.
At its core, these two models use past local weather station observations to forecast the weather variables of interest at different lead times into the future.
This structure is then extended with the gridded output of a large-scale weather model (could be NWP or ML) known to provide accurate forecasts on a broad scale, but lacking accuracy at fine scales (e.g., large-scale models largely neglect surface friction when modeling wind fields, see \Cref{fig:methods:data:era5vsmadis,fig:methods:data:era5vsmadis_distribution}).
By integrating large-scale forecasts with localized weather data, we can view the task as a correction of large-scale forecasts rather than forecasting \emph{de novo}; 
that is, our model aims to correct the large-scale forecast toward the local reality based on prior local observations. 
This setup enables our model to achieve accurate off-grid near-surface weather forecasting.

\subsection{Model}
\begin{figure}[ht]
\begin{center}
        \includegraphics[width=0.7\linewidth]{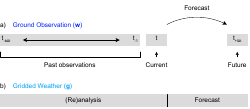}
    \end{center}
\caption{\textbf{Schematic of the forecasting task.} (a) The ground stations' past observations $\mathbf{w}(t-b \Delta t:t)$ are used to forecast the weather conditions at a given lead time $\mathbf{w}(t+l\Delta t)$. 
    (b) By introducing a gridded weather model's past and future data $\mathbf{g}$, the setup is transformed from a pure forecasting problem to a correction problem, where the future gridded weather data are corrected towards local observations.}
    \label{fig:methods:forecastingscheme}
\end{figure}

    The fundamental idea of weather forecasting is to predict the weather at a future time $l \Delta t$ (the lead time), given a set of information:
    \begin{equation}
        \mathbf{w}(t+l \Delta t) = F(\ldots),
    \end{equation}
    \noindent where $t$ is the current time, $\mathbf{w}$ is a vector of weather observations at $n$ different weather stations ($\mathbf{w} = [w_0, \ldots, w_n]$), 
    and $F$ is the function mapping input variables to the forecast.
    When using local historical data to predict the weather, the function $F$ takes the form:
    \begin{equation}
        \mathbf{w}(t+l \Delta t) = F(\mathbf{w}(t-b \Delta t:t)),
    \end{equation}
    \noindent where $b \Delta t$ is the number of past time steps considered, called back hours.
    This equation thus maps past weather data to future weather data, only considering the local weather stations (\Cref{fig:methods:forecastingscheme}a).

    We propose to change the nature of the problem, transforming the arguably hard task of forecasting to correcting an existing weather forecast.
    We thus introduce an external global weather forecast $\mathbf{g}$, and modify the function $F$:
    \begin{equation}
        \mathbf{w}(t+l \Delta t) = F(\mathbf{w}(t-b \Delta t:t), \mathbf{g}(t-b \Delta t:t + l \Delta t)).
    \end{equation}
    
    The global weather forecast covers the period from the back hours all the way to the lead time (\Cref{fig:methods:forecastingscheme}b).
    The function $F$ can take the form of any model, for instance a transformer or a GNN, which considers spatial correlation, i.e. the connections between the weather stations, in addition to temporal correlation.

    \subsubsection{Transformer}
        We implement a transformer model for the prediction function $F$.
        The transformer model is an encoder-only architecture similar to vision transformers~\cite{dosovitskiy2021a}.
        It treats weather measurements at each station as a token, then applies a series of self-attention layers to the tokens to learn spatial dependencies.
        Given the relative small size of the study area (Northeastern US), the transformer ingests all available weather stations at once.
        To integrate the gridded large-scale weather data, each station token is concatenated with the large-scale weather data at the grid cell closest to the station.
        
\begin{figure}[htbp]
    \begin{center}
        \includegraphics[width=.9\linewidth]{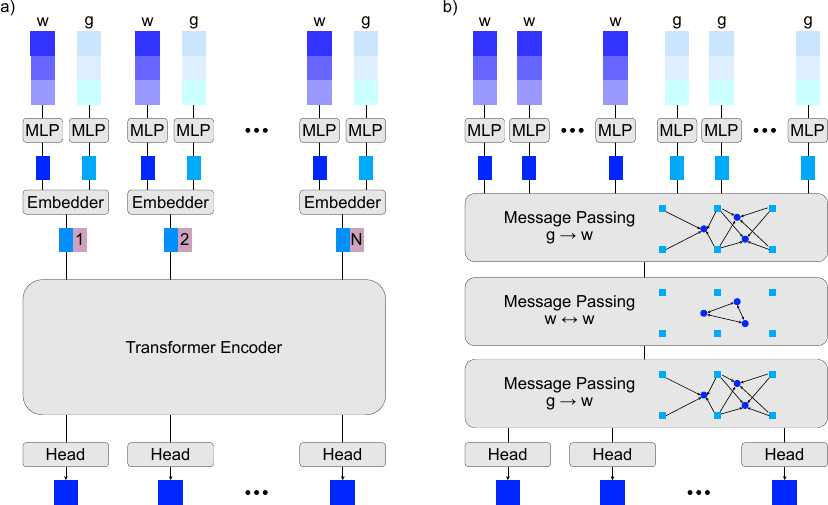}
    \end{center}
    \caption{\textbf{Transformer and GNN model for off-grid weather forecasting.}
    (a) Transformer Model: time series of gridded weather $\mathbf{g}$ and ground observation $\mathbf{w}$ at $N$ stations are combined using an MLP-based embedder to create token embeddings.
    These token embeddings are then concatenated with positional embeddings and passed through a transformer encoder, followed by an MLP prediction head, to generate forecasts for each station.
    (b) GNN Model: time series of gridded weather $\mathbf{g}$ and ground observation $\mathbf{w}$ at $N$ stations are embedded using an MLP embedder to create token embeddings for each modality separately.
    These token embeddings are then passed through three message passing network (MPNN) steps, first from global to local ($\mathbf{g}\rightarrow\mathbf{w}$), then local to local ($\mathbf{w}\leftrightarrow\mathbf{w}$), and finally from global to local again ($\mathbf{g}\rightarrow\mathbf{w}$).
    The MPNN is followed by an MLP prediction head, to generate forecasts for each station.
    }
    \label{fig:methods:modelStructure}
\end{figure}

        The overall structure of the transformer is given in \Cref{fig:methods:modelStructure} (a).
        Ground observation data $w$ at each station are combined with its nearest neighbor gridded weather data $g$ for $N$ stations, and
        passed through an MLP embedder to create token embeddings. 
        These token embeddings are then paired with positional embeddings and passed through a transformer encoder. 
        Finally, an MLP prediction head is applied to generate weather forecasts for each of the $N$ stations.

        Since the transformer encoder is the main component of the transformer model, we provide more details.
        For a complete description of the transformer encoder, please refer to the original paper~\cite{dosovitskiy2021a}.
        The encoder consists of a series of encoding blocks, each containing a multi-head self-attention mechanism and a feed-forward network.
        The multi-head self-attention mechanism allows the model to attend among input station tokens to learn representations that consider station-to-station spatial dependencies.
        The feed-forward network is a simple two-layer MLP aiming to refine the learned representations.

        \textbf{Multi-Head Self-Attention}

        In self-attention, each input token is first mapped to three different vector representations: \emph{query}, \emph{key}, and \emph{value}.
        Conceptually speaking, 
        the query vector represents what a token is looking for--its request for relevant information.
        The key vector represents the content or characteristics of a token--essentially what it can offer to other tokens.
        The value vector contains the actual information that will be passed along if a token is deemed relevant.
        These vectors are computed from the input tokens using learned linear projections, that is,
        \begin{equation}
        \mathbf{Q = XW_Q},  \mathbf{K = XW_K},  \mathbf{V = XW_V},
        \end{equation}
        where $\mathbf{W_Q}, \mathbf{W_K}, \mathbf{W_V}$ are the learned matrices transforming input token matrix $\mathbf{X} \in \mathbb{R}^{N \times d}$;
        and $\mathbf{Q}, \mathbf{K}, \mathbf{V} \in \mathbb{R}^{N \times d}$ are the matrices collecting query, key, and value vectors of each token.

        To determine how much attention each token should pay to every other token, 
        the model compares a token’s query with the keys of all tokens. 
        This is done via a scaled dot product:    
        \begin{equation}
            \text{Attention}(\mathbf{Q}, \mathbf{K}, \mathbf{V}) = \text{softmax}\left(\frac{\mathbf{Q}\mathbf{K}^T}{\sqrt{d}}\right)\mathbf{V}.
        \end{equation}
        The softmax converts these similarity scores from the dot product into a probability distribution, which is then used to compute a weighted sum of the value vectors. 
        This produces a new context-aware representation for each token, 
        where the contribution of other tokens is determined by their relevance to the current token’s query.

        In multi-head self-attention, this process is repeated $h$ times with different learned projections, and the results are concatenated and linearly transformed to produce the final output:
        \begin{equation}
            \text{MultiHead}(\mathbf{X}) = \text{Concat}(\text{head}_1, \ldots, \text{head}_h)\mathbf{W_O},
        \end{equation}
        where $\text{head}_i = \text{Attention}(\mathbf{XW_Q}^i, \mathbf{XW_K}^i, \mathbf{XW_V}^i)$ and $\mathbf{W_O} \in \mathbb{R}^{hd \times d}$ is the output projection matrix.

        \textbf{Feed-Forward Network and Residual Connections}

        Following the multi-head self-attention mechanism, each encoder block includes a feed-forward network (FFN) that is independently applied to each token. This component consists of two linear transformations with a non-linear activation in between, typically using the Rectified Linear Unit (ReLU) or GELU activation. Formally, given an input token representation $\mathbf{x} \in \mathbb{R}^d$, the FFN is defined as:

        \begin{equation}
            \text{FFN}(\mathbf{x}) = \max(0, \mathbf{x} \mathbf{W}_1 + \mathbf{b}_1) \mathbf{W}_2 + \mathbf{b}_2,
        \end{equation}
        where $\mathbf{W}_1 \in \mathbb{R}^{d \times d_h}$ and $\mathbf{W}_2 \in \mathbb{R}^{d_h \times d}$ are learned weights, and $d_h$ is the hidden dimension which is typically larger than $d$ (e.g., $d_h = 4d$). $\mathbf{b}_1 \in \mathbb{R}^{d_h}$ and $\mathbf{b}_1 \in \mathbb{R}^{d}$ are the biases within the FFN. To facilitate optimization and preserve gradient flow, each sub-layer (i.e., the self-attention and the FFN) is wrapped with a residual connection followed by layer normalization~\cite{ba2016}.

    \subsubsection{Message Passing Neural Network (MPNN)}
        As a baseline, we also implement the prediction function $F$ as a GNN. 
        The overall structure of the GNN is given in \Cref{fig:methods:modelStructure} (b).
        Each weather station naturally becomes a node of a graph. 
        The weather station graph is constructed with Delaunay triangulation~\cite{delaunay1934} on the geographic coordinates of the stations. Delaunay triangulation links each station to a few nearby neighbors, giving balanced connections in all directions while maintaining sparsity and efficient message-passing.
        
\begin{figure}[t]
    \begin{center}
        \includegraphics[width=.8\linewidth]{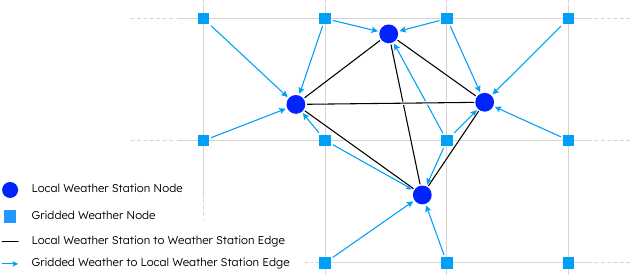}
    \end{center}
    \caption{
        \textbf{Simplified diagram of our multi-modal graph.}
        Local weather stations form the base graph, with each station node connected to its neighbors using Delaunay triangulation.
        The gridded numerical weather dataset is arranged on a regular mesh, with each station node connected to its $o$ closest gridded nodes (4 in this example). 
        Station nodes pass messages to each other in bi-directional edges; 
        grid nodes pass messages to station nodes, but not vice-versa.
    }\label{fig:methods:globalweathergraph}
\end{figure}

        To integrate the large-scale weather data (past and future), the weather station graph is extended to include NWP grid cell nodes (\Cref{fig:methods:globalweathergraph}).
        In this heterogeneous graph, each weather station node is connected to the $o$ closest large-scale weather data grid cell nodes (4 in the example in \Cref{fig:methods:globalweathergraph}, but 8 in the experiments later).
        These edges are uni-directional, meaning the information flows from global to local, but not back. 
        The heterogeneous graph constructed for our study area is given in \Cref{fig:methods:experiments}.
    
        MPNNs are a type of GNNs that operate on a given graph structure by passing messages between connected nodes.
        The messages consist of information contained in the nodes as well as in the edges connecting the nodes.
        The nodes are updated with the incoming messages.
        This architecture can be trained for different tasks, such as predicting at a node level (e.g., simulating particle dynamics) and at a graph level (e.g., classifying chemicals).
        We follow the implementation of MPNN as described in~\citeA{brandstetter2023}.
        It works in three steps: encode, process, and decode~\cite{battaglia2018, sanchez-gonzalez2020}.

        \textbf{Encode}
        This step encodes the information contained in each node and transforms it into a latent feature. 
        Station nodes and large-scale weather data grid cell nodes are encoded separately.
 
        \begin{equation}
            f_i \leftarrow \alpha(w_i(t-b \Delta t:t), p_i),
            \label{eq:methods:model:station_encoding} 
        \end{equation}

        \begin{equation}
            h_r \leftarrow \psi(g_r(t-b \Delta t:t+l \Delta t), p_r),
            \label{eq:methods:model:grid_encoding}
        \end{equation}
        
        \noindent where $w_i$ is a vector containing the observed weather variables at station node $i$; 
        $g_r$ is the large-scale weather data at grid cell node $r$;
        $p$ the coordinates;
        and $\alpha, \psi$ are encoding neural networks, here a simple two-layer MLP.
        The $f, h$ denote embedded features at each node; they extract the essential information contained in station nodes and grid cell nodes separately.

        \textbf{Process} 
        This step processes each node's feature with incoming messages that aggregate information from its connected neighbors.
        As we are interested in the weather at the station level, only the station node features are updated.
        Since each station node has two types of neighbors, (1) other station nodes and (2) large-scale weather data grid cell nodes, the update procedure has two forms as well.
        First, update station $i$ with its station neighbors $j$:

        \begin{equation}
            \mu_{ij} \leftarrow \beta(f_i, f_j, w_i(t-b \Delta t:t) - w_j(t-b \Delta t:t), p_i - p_j),
            \label{eq:methods:model:mpnn:message}
        \end{equation}

        \begin{equation}
            f_i \leftarrow f_i + \gamma \left( f_i,  \frac{1}{\vert \mathcal{N}(i) \vert} \sum\limits_{j\in\mathcal{N}(i)}{\mu_{ij}} \right)
            \label{eq:methods:model:mpnn:update}
        \end{equation}

        \noindent $\mu_{ij}$ is the produced message passing from station node $j$ to station node $i$. 
        Station node feature $f_i$ is then updated according to eq. \ref{eq:methods:model:mpnn:update} with the message from its station neighbors whose indices are in $\mathcal{N}(i)$.
        $\beta$ and $\gamma$ are two-layer MLPs completing the process.

        Second, update with the large-scale weather data grid cell node neighbors:
        \begin{equation}
            \nu_{ir} \leftarrow \chi(h_r, f_i, p_i - p_r)
            \label{eq:methods:model:mpnn:message_external}
        \end{equation}

        \begin{equation}
            f_i \leftarrow f_i + \omega\left( f_i, h_r,  \frac{1}{\vert \mathcal{M}(i) \vert} \sum\limits_{r\in\mathcal{M}(i)}{\nu_{ir}} \right),
            \label{eq:methods:model:mpnn:update_external}
        \end{equation}
        
        \noindent Similarly, $\nu_{ir}$ is the message passing information at weather data grid cell node $r$ to station node $i$.
        Station node feature $f_i$ is updated with message $\nu_{ir}$ by eq. \ref{eq:methods:model:mpnn:update_external}, where
        $\mathcal{M}(i)$ contains all cell nodes connected to station node $i$. 
        $\chi$ and $\omega$ are two-layer MLPs containing all trainable parameters within the process.
        
        \textbf{Decode}
        The decoding step then maps the final station node feature to the weather variables at the given lead time:
        
        \begin{equation}
            w_i(t + l\Delta t) = \phi(f_i),
            \label{eq:methods:model:localdecoding}
        \end{equation}
        \noindent with $\phi$ a two-layer MLP.

        \textbf{Predictive Graph Model}
        Putting everything together, weather variables at all stations are predicted by:
        \begin{equation}
            \mathbf{w}(t + l\Delta t) = \mathbf{D} \circ \hat{\mathbf{P}}_2 \circ \cdots \circ \mathbf{P}_k \circ \cdots \circ \hat{\mathbf{P}}_1 \circ \mathbf{E}~(G). 
        \end{equation}

        \noindent The multi-modal input data is organized as a graph $G$. 
        Its nodes are first encoded by $\mathbf{E}$. 
        Station nodes are then updated with information from grid cell nodes by $\hat{\mathbf{P}}_1$.
        After the grid node message passing, the station nodes are updated internally by multiple layers of station node message passing $\mathbf{P}_k$.
        Following that, the station nodes ingest grid cell node information one more time by $\hat{\mathbf{P}}_2$.
        Finally, the station nodes are decoded by $\mathbf{D}$ to produce the weather variables of interest.
    
\subsection{Data}
Our goal in this work is to forecast the weather at precise locations, integrating historical observations, and supplemented with global gridded weather products. 
To do this, we prepare following datasets: 
(1) point-based weather observations from MADIS stations,
(2) gridded reanalysis data from ERA5, and
(3) gridded weather forecast products from HRRR.
The details of our curated multi-modal dataset are summarized in \Cref{tab:methods:dataset}.
ERA5 reanalysis has lower resolution and is less optimized for US weather modeling than HRRR, but is widely used as ground-truth for foundation models and is thus a good candidate for a baseline forecast. 
Additionally, performing experiments with both allows us to study how the quality of the global weather data affects weather forecasting at station level.

\textbf{MADIS}
The Meteorological Assimilation Data Ingest System 
(MADIS)
is a database provided by the National Oceanic and Atmospheric Administration (NOAA) that contains meteorological observations from stations covering the entire globe. 
MADIS ingests data from NOAA and non-NOAA sources, including observation networks from US federal, state, and transportation agencies, universities, volunteer networks, and data from private sectors like airlines as well as public-private partnerships like the Citizen Weather Observer Program. 
MADIS provides a wide range of weather variables from which we curated 10m wind speed, 10m wind direction, 2m temperature, and 2m dewpoint temperature for this study.
In this work, we focus on stations over the Northeastern US region (Maine, New Hampshire, Vermont, Massachusetts, Rhode Island, Connecticut, New York, New Jersey, and Pennsylvania, see \Cref{fig:methods:data:era5vsmadis}a). 
We only keep averaged hourly observations with the quality flag ``Screened'' or ``Verified'',
which are annotated following ~\citeA{ramavajjala2023}.
Additionally, only stations with at least 90\% of data of sufficient quality over the 5 years are considered.
We front filled the data for the rare stations with missing data (10\% for the worst stations).
Across the study region, this leaves us with 358 stations (\Cref{fig:methods:data:era5vsmadis}a, dark blue arrows). 
We processed 5 years of data from 2019 to 2023.

\textbf{ERA5}
The ECMWF Reanalysis v5 (ERA5) climate and weather dataset~\cite{hersbach2020} is a gridded reanalysis product from the European Center for Medium-Range Weather Forecasts (ECMWF) that combines model data with worldwide observations. 
The observations are used as boundary conditions for numerical models that then predict various atmospheric variables. 
ERA5 is available as global hourly data with a $0.25^{\circ}\times0.25^{\circ}$ resolution, which is $31~\mbox{km}/\text{pixel}$ at the equator, spanning 1950--2024. 
It includes weather both at the surface and at various pressure levels. 
We curated 5 years (from 2019 to 2023) of surface variables: 10m wind $u$, 10m wind $v$, 2m temperature, and 2m dewpoint temperature.

\textbf{HRRR}
The High-Resolution Rapid Refresh (HRRR, \citeA{dowell2022}) model, developed by NOAA, is a numerical weather prediction system that provides high-resolution, short-term weather forecasts for the continental US and Alaska. 
Designed for detailed and timely predictions, HRRR operates with a grid spacing of approximately 3 kilometers and updates hourly, incorporating real-time radar, satellite, and observational data. 
Each update generates forecasts up to 18 hours ahead,
along with an analysis of the current atmospheric state (considered ground truth). 
Note that forecasts produced at 00, 06, 12, 18 UTC provide data up to 48 hours lead times.
However, for uniformity, only forecasts up to 18 hours are used here.
For this study, we curated 5 years (2019--2023) of surface variables from HRRR, including 10m wind components ($u$ and $v$), 2m temperature, and 2m dewpoint temperature.
    
\subsection{Experiments}\label{sec:methods:experiments}

    \textbf{Forecast Setup} 
    In this work, models are trained to predict 10m wind vector ($u$ and $v$), 2m temperature, and 2m dewpoint temperature all together at each MADIS station.
    To predict these variables at each weather station, we first provide the model with prior measurements of 10m wind $u$, 10m wind $v$, 2m temperature, and 2m dewpoint temperature at each MADIS node. 
    Similarly, at each grid cell of ERA5 or HRRR, the inputs are 10m $u$, 10m $v$, 2m temperature, and 2m dewpoint temperature (from past and future).
    For all inputs, the temporal resolution is 1 hour.
    The transformer model is given all weather stations as a sequence of tokens, and each station token is concatenated with the nearest NWP grid cell.
    In the MPNN graph, each MADIS node is connected to its MADIS neighbors based on the Delaunay triangulation.
    Then each MADIS node is also connected to its 8 nearest NWP grid cells based on the Euclidean distance (\Cref{fig:methods:experiments}).
    
    All models are tasked with predicting 10m $u$, 10m $v$, 2m temperature, and 2m dewpoint temperature at each MADIS node for different lead times: 1, 2, 4, 8, 12, 18, 24, 36 and 48 hours, for which we train one model each.
    The model is trained with data from 2019 to 2021, validated on 2022, and tested on 2023.
    There are $\sim$8,760 time steps per year for each of the 358 stations.
    In total, we obtain a training set of $\sim$26,280 samples, a validation set of $\sim$8,760 samples, and a test set of $\sim$8,760 samples.
    The model uses 48 hours of MADIS back hours (\Cref{fig:methods:forecastingscheme}a), i.e. the weather observations from the previous 48 hours, including the current observation, to predict forward.
    When including large-scale gridded weather data, the model is given the time steps from the back hours to the lead time (\Cref{fig:methods:forecastingscheme}b), providing a full temporal view of large scale dynamics.
    Since the HRRR forecast is limited to 18 hours ahead, models use HRRR forecast data up to 18 hours for lead times beyond 18 hours, resulting in a shorter time series of forecast data as input. 
    
    \textbf{Baseline Methods} 
    Besides the MPNN model, we also compare the transformer against a series of other baseline forecasting methods: interpolated ERA5, interpolated HRRR (including both analysis up the current time and forecasts into the future), MADIS persistence, and an MLP.
    Interpolated ERA5 or HRRR refers to a nearest neighbor interpolation of ERA5 or HRRR grid cells to a MADIS station location,
    meaning each station location takes ERA5 or HRRR value at the grid cell closest to the station.
    The MADIS persistence simply shifts the observation by the lead time and will perform well if the temporal auto-correlation of weather variables is high.
    The MLP provides a baseline model with an architecture mirroring the architecture of the MPNN, 
    but with no station-to-station message passing spatial structure.
    For the MLP experiments, the gridded weather data is interpolated at the weather stations and used as an additional input. 
    The same MLP is tasked with forecasting at all stations; 
    we also tried training a separate MLP for each station but did not observe better performance.
    Note that the MLP uses the same architecture as the MPNN, but without the message passing structure and inter-station connections.
    The nearest embedded external node is concatenated with the station token to replace the message passing from the MPNN.

    \textbf{Study of Gridded Weather Product Contributions}
    The MLP, MPNN, and transformer are run with and without gridded weather data to assess how much performance gain for localized weather forecasting comes from knowing large-scale weather dynamics.
    When including gridded weather data, models are run with ERA5 reanalysis, HRRR forecast or with HRRR analysis.
    Both HRRR forecast and analysis are used to assess how the error in large-scale weather forecasts are propagated to the localized weather forecasting.
    In addition, ERA5 has lower resolution than HRRR; 
    HRRR is optimized for the US while ERA5 is optimized for the globe.
    To see how this data quality difference affects weather forecasting at the station level, models run with ERA5 reanalysis are also compared to those run with HRRR analysis.

\section{Results}
\begin{figure}[htbp]
    \begin{center}
\includegraphics[width=\linewidth]{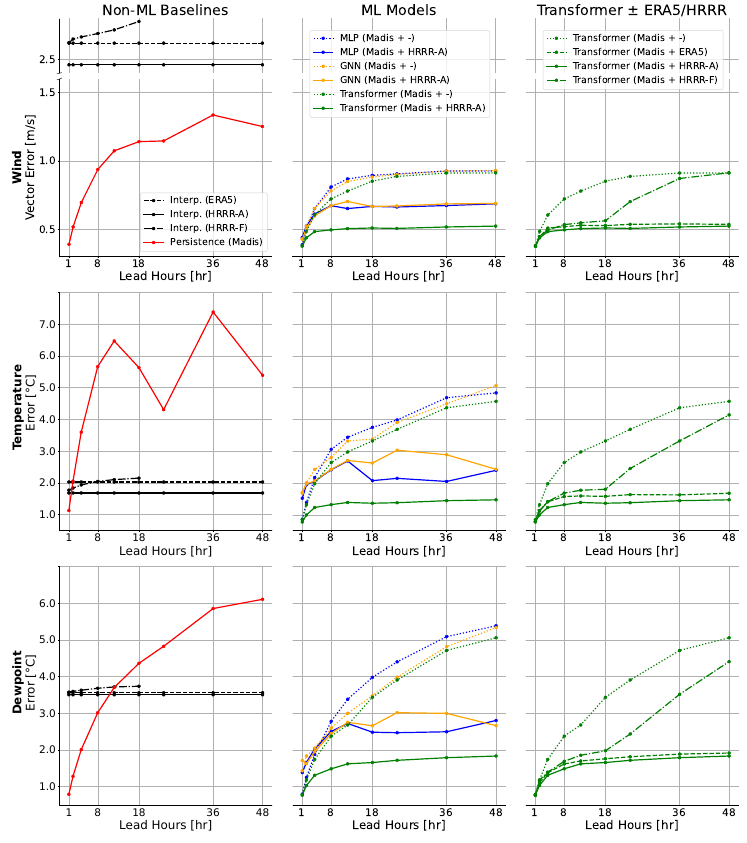}
    \end{center}
    \caption{
        \textbf{Model forcast errors with respect to 10m wind, 2m temperature, and 2m dewpoint at various lead times.} 
        From top to bottom: model performance on 10m wind, 2m temperature, and 2m dewpoint is shown as a function of lead time.
        The left column shows naive non-ML model performance, the middle column shows ML model performance with/without gridded weather data, and the right column shows model performance of only transformer models but with different types of gridded weather data as inputs.
        Please refer to \Cref{tab:results:Wind}, \ref{tab:results:Temp}, and \ref{tab:results:Dew} for detailed numerical results, as well as \Cref{tab:ablation:structure} for an extended ablation study on model structure.
    }
    \label{fig:results:modelerrors}
\end{figure}

We report the model performance on the 2023 test set for 10m wind, 2m temperature, and 2m dewpoint separately.
The metric used for 2m temperature and dewpoint is the standard root mean square error (RMSE).
As for 10m wind (a vector containing $u$ and $v$ components), 
the metric used is the mean magnitude of the wind vector error ($\sqrt{\Delta u^2 + \Delta v^2}$).
\Cref{fig:results:modelerrors} summarizes forecast performance of different models at various lead times with respect to all three variables.
Detailed numerical results are provided in \Cref{tab:results:Wind}, \ref{tab:results:Temp}, and \ref{tab:results:Dew}.

\textbf{Transformer is the best forecasting model for all variables}
The second column of \Cref{fig:results:modelerrors} shows that the transformer model outperforms the GNN and MLP models for all variables whether gridded HRRR analysis data is used or not.
When gridded HRRR analysis data is used, the transformer model also performs better than the non-ML models 
such as the persistence model and the gridded weather interpolation models for all variables as demonstrated in the first column of \Cref{fig:results:modelerrors}.
Specifically, for 10m wind, the transformer model with HRRR-A (HRRR analysis data) achieves a lead-hour averaged wind vector error of 0.48 m/s, 
which is 22\% lower than the next best performing ML model, the MLP model with HRRR-A. 
It is also 49\% lower than the persistence model and 80\% lower than the HRRR-A interpolation models.
In terms of 2m temperature, the transformer model with HRRR-A reduces the lead-hour averaged RMSE by 41\% compared to the MLP model with HRRR-A. 
 Its RMSE is also 73\% lower than the persistence model and 25\% lower than the HRRR-A interpolation model.
In addition, the transformer model with HRRR-A achieves a lead-hour averaged 2m dewpoint RMSE of 1.46 Celsius degrees,  which is 36\% lower than the MLP model with HRRR-A, 59\% lower than the persistence model, and 58\% lower than the HRRR-A interpolation model.
\begin{figure}[htbp]
    \begin{center}
\includegraphics[width=\linewidth]{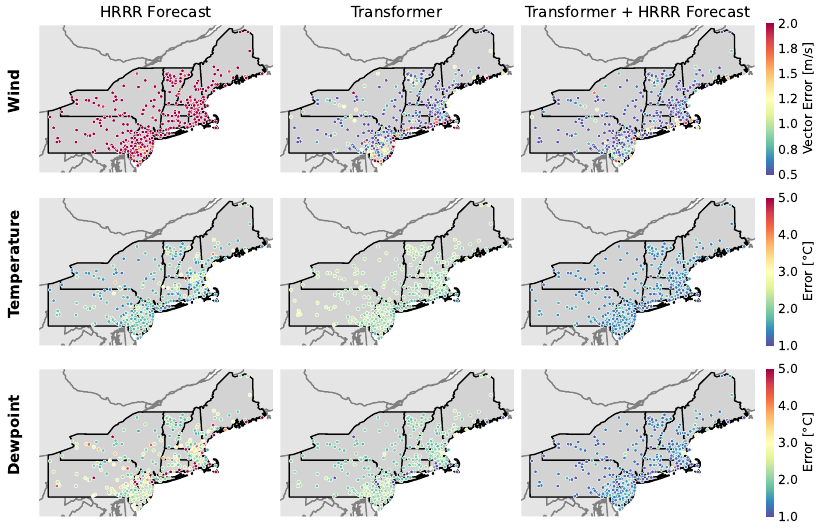}
    \end{center}
    \caption{
        \textbf{Forecast errors of the different methods shown for each MADIS station, averaged over the lead times less than 18 hours.}
        From top to bottom: forecast error spatial distribution is presented for 10m wind, 2m temperature, and 2m dewpoint.
        The transformer models with HRRR forecast (third column) show significantly lower errors than the naive interpolations of HRRR forecast (first column) and the transformer models with only MADIS input (second column).
    }\label{fig:results:mse_spatial}
    \vspace{-.7cm}
\end{figure} 
\textbf{Gridded weather data is beneficial for all ML models}
The second column of \Cref{fig:results:modelerrors} shows that all ML models perform better when gridded HRRR analysis data is used.
This performance boost is also true across all variables.
For example, the HRRR-A data helps the transformer model reduce lead-hour averaged wind vector error by 33\%,
temperature RMSE by 56\%, and dewpoint RMSE by 49\%, compared to the counterparts without HRRR-A.

\textbf{The quality of gridded weather data is critical for ML models}
The quality of gridded weather data is directly reflected in the performance of resultant ML models who use it.
As shown in the first column of \Cref{fig:results:modelerrors}, HRRR-A interpolation outperforms HRRR-F (HRRR forecast) 
interpolation; and ERA5 interpolation achieves performance in between the two.
Here HRRR-A(ERA5) interpolation uses future (re)analysis as forecasts.
As a result, transformer models with gridded weather data share the same order of performance: with HRRR-A better than with ERA5, and with ERA5 better than with HRRR-F
(the third column of \Cref{fig:results:modelerrors}).
Still, all transformer models with gridded weather data outperform the one without, as gridded data provides a future outlook of larger scale weather patterns. 
Additionally, there is a noticeable increase in errors for transformer models with HRRR-F beyond the 18 hour lead time, 
which aligns with the fact that HRRR only provides forecast data up to 18 hours ahead.

\textbf{Wind prediction benefits the most from combining gridded weather data with station data}
The transformer model combining HRRR-A and station measurements reduces wind vector error by 80\%, temperature RMSE by 25\%, and dewpoint RMSE by 58\%, 
compared to naive HRRR-A interpolation.
The wind vector error is the most reduced, followed by dewpoint RMSE, and temperature RMSE is the least reduced.
We speculate that local wind is the variable that is the most sensitive to local environment and/or modeled the most poorly by global gridded models, followed by dewpoint and temperature.
This is also observed in \Cref{fig:methods:data:era5vsmadis,fig:methods:data:era5vsmadis_distribution}: temperature and dewpoint ERA5 reanalysis data correlate well with the station data, 
while wind has much lower correlation.
\begin{figure}[htbp]
    \begin{center}
        \includegraphics[width=1\linewidth]{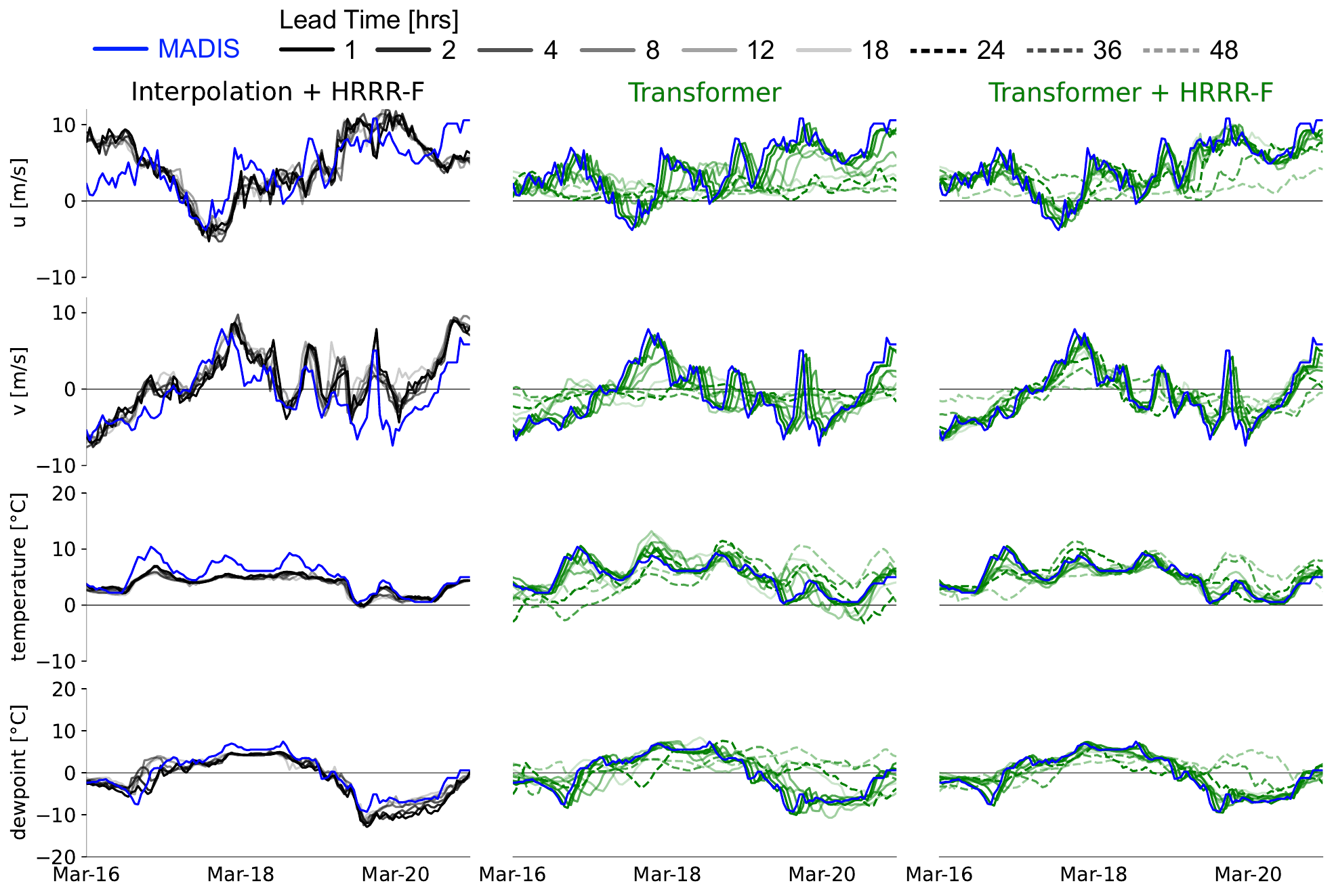}
    \end{center}
    \caption{
        \textbf{Example time series illustrating the transformer model's integration of HRRR-F into its prediction.}
        From top to bottom: time series of 10m $u$ and $v$ components, 2m temperature, and 2m dewpoint are presented.
        From left to right: HRRR forecast interpolation, transformer with MADIS only, and transformer with additional HRRR forecasts are demonstrated.
        Each panel has the MADIS ground truth in blue, and the predictions at increasing lead times displayed with decreasing saturation.
        This station happens to be the one where the transformer + HRRR-F model performs the worst on average. For this station and time snippet the interpolated HRRR-F appears relatively accurate, and the transformer is able to integrate the HRRR-F data well.
    }\label{fig:results:time_series_T_HRRR_F_bad}
    \vspace{-.7cm}
\end{figure} 
\textbf{Transformer model with HRRR forecast improves local weather predictions across all stations}
\Cref{fig:results:mse_spatial} illustrates how prediction errors from the transformer models vary across different stations.
Compared to the naive interpolations of HRRR forecast (first column), the transformer model informed by only station measurements (second column) show lower errors for 10m wind and 2m dewpoint, but slightly higher errors for 2m temperature.
This indicates that wind and dewpoint are more sensitive to local environments; while temperature is more dominated by large scale atmospheric conditions.
When using HRRR-F (third column), the transformer model's performance is further improved, especially for 2m temperature and 2m dewpoint.
The resultant tranformer model shows significantly lower errors than the naive interpolations of HRRR-F for all three variables. 
This error reduction is seen across all stations, indicating that the transformer model is able to generalize well across different locations.
Interestingly, the transformer model shows relatively large wind errors at stations along the coastline. It may be due to the fact that these stations are situated in an open environment, thus experiencing similar conditions represented by gridded weather models. As a results, they benefit less from local measurements.

\textbf{Transformer model integrates HRRR forecast data well}
\Cref{fig:results:time_series_T_HRRR_F_bad} shows, for different models, a snippet of time series of forecasted 10m wind $u$ and $v$ components, 2m temperature, and 2m dewpoint at the station with the worst forecast performance for the transformer model with HRRR-F.
Unlike at many other stations, naive HRRR-F interpolation for this location agrees with the local MADIS data well, especially for 10m wind components and 2m temperature (c.f. \Cref{fig:results:time_series_T_HRRR_F_good} for a time series where the naive interpolation doesn't follow the MADIS observation as well).
This allows us to evaluate the transformer model's ability to integrate HRRR forecast data.
The transformer model integrates the HRRR forecast data well, consistently correcting it towards the local measurements, and ignoring it when needed (c.f. wind in \Cref{fig:results:time_series_T_HRRR_F_good}).
The further out the lead time, the more it relies on the HRRR-F data, but still performs better than the interpolated HRRR-F.
Meanwhile, without HRRR-F, the tranformer model fails for longer lead times; for instance, it trends toward predicting a flat wind vector, preferring to minimize error by predicting the average. 
\section{Discussion}
\textbf{Weather station observations and gridded weather data are complementary for successful local weather forecasting}
Discrepancies arise between weather station observations and gridded weather data, such as ERA5 and HRRR, due to the latter's limitations in modeling terrain and land cover. 
We have shown that these local surface characteristics significantly influence local weather patterns, particularly wind (\Cref{fig:methods:data:era5vsmadis}).
Meanwhile, the high forecast error associated with MADIS persistence shows that current local weather conditions do not reliably indicate future conditions. 
Accurate fine-grained weather forecasting therefore requires a model that accounts for both local conditions and the inherent variability of local weather patterns across time.
Gridded weather products and historical station data provide key complementary information for capturing local weather dynamics. 
By combining these two datasets, we achieve a local weather forecasting model that is significantly more accurate than ERA5 or HRRR (and, by extension, current gridded ML weather models). 
The need for weather station data to capture local conditions is especially evident where stations situated amidst buildings or trees experience conditions that differ significantly from those represented by gridded weather models (\Cref{fig:appendix:stations_comparison}).
This is most apparent inlands, where the reduction in error of the gridded dataset is greatest, but also on the coast for more sheltered weather stations.

\textbf{The dynamic spatial attention mechanism gives the transformer an edge in local weather forecasting}
In this work, we explore three ML models: a transformer, a GNN, and an MLP, 
that can integrate both local weather station observations and gridded weather data successfully for local weather forecasting. 
When predicting local weather using only local data, all ML models achieve respectable performances. 
With the introduction of ERA5 or HRRR, all models successfully incorporate large-scale atmospheric dynamics from ERA5 or HRRR, 
thereby improving predictions at longer lead times.
Among the three models, the transformer model shows the best performance.
This is likely due to the transformer's dynamic spatial attention mechanism.
It enables the transformer to attend to the relevant stations according to current weather conditions.
In contrast, the GNN can only attend to predefined station locations, and
the MLP is unable to attend to any other stations besides each station itself (see \Cref{tab:ablation:structure} for an extended ablation study on model structure).

\textbf{Direct input of station data helps make predictions at station locations more accurate}
Using both station data and gridded weather data for local weather forecasting is not new.
For example, ~\citeA{allen2025} trained an end-to-end ML local weather forecasting model 
that first assimilates station data into gridded weather data;
then makes time stepping into the future in the gridded space;
lastly, downscales the gridded weather forecast to the station locations.
However, their model's performance is not as good as the transformer model in this work, especially for surface wind predictions, where their wind prediction error caps out at 1 m/s, while the transformer model in this work gets it down to 0.5 m/s.
This significant performance gap may be attributed to the fact that 
when making final predictions at station locations,
their model only uses the gridded weather forecast as input but ignores the station data used to assimilate the gridded fields.
In other words, their model lacks the direct input of station data when making station-level predictions.
In contrast, the transformer model in this work directly uses the station data as input and combines with gridded data when making predictions.
It gives the transformer model a better understanding of the station location weather patterns,
which is crucial for local weather forecasting.
In summary, our approach provides a phase shift in local weather forecasting accuracy when directly combining station data with gridded weather data.
All the ML based gridded weather forecasting models, whether foundation models or end-to-end models,
will be limited in local weather forecasting accuracy without the direct input of station data.

\section{Conclusion}
This work demonstrates the use of a multi-modal transformer for downscaling gridded weather forecasts and improving the accuracy of off-grid predictions. 
Our model addresses the inherent bias in gridded numerical weather products like ERA5 and HRRR. 
By concatenating both MADIS weather station data and HRRR or ERA5 gridded weather data as input tokens,
our transformer predicts off-grid weather conditions by leveraging both large-scale atmospheric dynamics and local weather patterns.
In our evaluation of surface variable (10m wind, 2m temperature, and 2m dewpoint) prediction tasks, the transformer outperforms all baseline models. 
For instance, it achieved a 58\% reduction in overall prediction error compared to HRRR renalysis interpolation and a 36\% improvement over the best-performing ML model. 
An ablation study, where numerical weather inputs were removed, all ML models performed significantly worse, highlighting the importance of incorporating large-scale atmospheric dynamics for accurate local predictions.
This finding motivates the exploration of additional modalities, such as radar measurements and satellite imagery, which could further enhance local forecast accuracy.
This research has significant implications for improving weather forecasting, particularly in high-value regions where weather stations can be installed.
More accurate off-grid predictions can enhance weather-dependent decision-making in various sectors, including agriculture, wildfire management, transportation, and renewable energy.  
Future work will focus on expanding the study area and exploring the integration of our transformer model with weather foundation models.

\vspace{.5cm}
\noindent\textbf{Code and Data Availability}\\
The code and data for this paper are available, on GitHub (\url{https://github.com/Earth-Intelligence-Lab/LocalizedWeather/}) and Zenodo~\cite{yang2025Dataset, yang2025Code}.

\acknowledgments
We would like to thank and acknowledge Shell Information Technology International Inc. for funding the project, Peetak Mitra for helping processing the MADIS data, and our partners at Amazon Web Services and its Solutions Architects, Brian McCarthy, Jared Novotny and Dr. Jianjun Xu, for providing cloud computing and AWS technical support.

\bibliography{references_jg}

\begin{thebibliography}{}

\bibitem [\protect \citeauthoryear {%
Allen%
\ \protect \BOthers {.}}{%
Allen%
\ \protect \BOthers {.}}{%
{\protect \APACyear {2025}}%
}]{%
allen2025}
\APACinsertmetastar {%
allen2025}%
\begin{APACrefauthors}%
Allen, A.%
, Markou, S.%
, Tebbutt, W.%
, Requeima, J.%
, Bruinsma, W\BPBI P.%
, Andersson, T\BPBI R.%
\BDBL {}Turner, R\BPBI E.%
\end{APACrefauthors}%
\unskip\
\newblock
\APACrefYearMonthDay{2025}{{\APACmonth{03}}}{}.
\newblock
{\BBOQ}\APACrefatitle {End-to-end data-driven weather prediction} {End-to-end data-driven weather prediction}.{\BBCQ}
\newblock
\APACjournalVolNumPages{Nature}{}{}{}.
\newblock
\begin{APACrefURL} [{2025-04-01}]\url{https://www.nature.com/articles/s41586-025-08897-0} \end{APACrefURL}
\newblock
\begin{APACrefDOI} \doi{10.1038/s41586-025-08897-0} \end{APACrefDOI}
\PrintBackRefs{\CurrentBib}

\bibitem [\protect \citeauthoryear {%
Andrychowicz%
\ \protect \BOthers {.}}{%
Andrychowicz%
\ \protect \BOthers {.}}{%
{\protect \APACyear {2023}}%
}]{%
andrychowicz2023}
\APACinsertmetastar {%
andrychowicz2023}%
\begin{APACrefauthors}%
Andrychowicz, M.%
, Espeholt, L.%
, Li, D.%
, Merchant, S.%
, Merose, A.%
, Zyda, F.%
\BDBL {}Kalchbrenner, N.%
\end{APACrefauthors}%
\unskip\
\newblock
\APACrefYearMonthDay{2023}{{\APACmonth{07}}}{}.
\newblock
\APACrefbtitle {Deep {Learning} for {Day} {Forecasts} from {Sparse} {Observations}.} {Deep {Learning} for {Day} {Forecasts} from {Sparse} {Observations}.}
\newblock
\APACaddressPublisher{}{arXiv}.
\newblock
\begin{APACrefURL} [{2023-11-02}]\url{http://arxiv.org/abs/2306.06079} \end{APACrefURL}
\newblock
\APACrefnote{arXiv:2306.06079 [physics]}
\newblock
\begin{APACrefDOI} \doi{10.48550/arXiv.2306.06079} \end{APACrefDOI}
\PrintBackRefs{\CurrentBib}

\bibitem [\protect \citeauthoryear {%
Ba%
, Kiros%
\BCBL {}\ \BBA {} Hinton%
}{%
Ba%
\ \protect \BOthers {.}}{%
{\protect \APACyear {2016}}%
}]{%
ba2016}
\APACinsertmetastar {%
ba2016}%
\begin{APACrefauthors}%
Ba, J\BPBI L.%
, Kiros, J\BPBI R.%
\BCBL {}\ \BBA {} Hinton, G\BPBI E.%
\end{APACrefauthors}%
\unskip\
\newblock
\APACrefYearMonthDay{2016}{{\APACmonth{07}}}{}.
\newblock
\APACrefbtitle {Layer {Normalization}.} {Layer {Normalization}.}
\newblock
\APACaddressPublisher{}{arXiv}.
\newblock
\begin{APACrefURL} [{2025-04-15}]\url{http://arxiv.org/abs/1607.06450} \end{APACrefURL}
\newblock
\APACrefnote{arXiv:1607.06450 [stat]}
\newblock
\begin{APACrefDOI} \doi{10.48550/arXiv.1607.06450} \end{APACrefDOI}
\PrintBackRefs{\CurrentBib}

\bibitem [\protect \citeauthoryear {%
Battaglia%
\ \protect \BOthers {.}}{%
Battaglia%
\ \protect \BOthers {.}}{%
{\protect \APACyear {2018}}%
}]{%
battaglia2018}
\APACinsertmetastar {%
battaglia2018}%
\begin{APACrefauthors}%
Battaglia, P\BPBI W.%
, Hamrick, J\BPBI B.%
, Bapst, V.%
, Sanchez-Gonzalez, A.%
, Zambaldi, V.%
, Malinowski, M.%
\BDBL {}Pascanu, R.%
\end{APACrefauthors}%
\unskip\
\newblock
\APACrefYearMonthDay{2018}{{\APACmonth{10}}}{}.
\newblock
\APACrefbtitle {Relational inductive biases, deep learning, and graph networks.} {Relational inductive biases, deep learning, and graph networks.}
\newblock
\APACaddressPublisher{}{arXiv}.
\newblock
\begin{APACrefURL} [{2024-04-17}]\url{http://arxiv.org/abs/1806.01261} \end{APACrefURL}
\newblock
\APACrefnote{arXiv:1806.01261 [cs, stat]}
\PrintBackRefs{\CurrentBib}

\bibitem [\protect \citeauthoryear {%
Bauer%
, Thorpe%
\BCBL {}\ \BBA {} Brunet%
}{%
Bauer%
\ \protect \BOthers {.}}{%
{\protect \APACyear {2015}}%
}]{%
bauer2015}
\APACinsertmetastar {%
bauer2015}%
\begin{APACrefauthors}%
Bauer, P.%
, Thorpe, A.%
\BCBL {}\ \BBA {} Brunet, G.%
\end{APACrefauthors}%
\unskip\
\newblock
\APACrefYearMonthDay{2015}{{\APACmonth{09}}}{}.
\newblock
{\BBOQ}\APACrefatitle {The quiet revolution of numerical weather prediction} {The quiet revolution of numerical weather prediction}.{\BBCQ}
\newblock
\APACjournalVolNumPages{Nature}{525}{7567}{47--55}.
\newblock
\begin{APACrefURL} [{2024-09-30}]\url{https://www.nature.com/articles/nature14956} \end{APACrefURL}
\newblock
\begin{APACrefDOI} \doi{10.1038/nature14956} \end{APACrefDOI}
\PrintBackRefs{\CurrentBib}

\bibitem [\protect \citeauthoryear {%
Bentsen%
, Warakagoda%
, Stenbro%
\BCBL {}\ \BBA {} Engelstad%
}{%
Bentsen%
\ \protect \BOthers {.}}{%
{\protect \APACyear {2023}}%
}]{%
bentsen2023}
\APACinsertmetastar {%
bentsen2023}%
\begin{APACrefauthors}%
Bentsen, L\BPBI {\textbackslash}.%
, Warakagoda, N\BPBI D.%
, Stenbro, R.%
\BCBL {}\ \BBA {} Engelstad, P.%
\end{APACrefauthors}%
\unskip\
\newblock
\APACrefYearMonthDay{2023}{{\APACmonth{03}}}{}.
\newblock
{\BBOQ}\APACrefatitle {Spatio-temporal wind speed forecasting using graph networks and novel {Transformer} architectures} {Spatio-temporal wind speed forecasting using graph networks and novel {Transformer} architectures}.{\BBCQ}
\newblock
\APACjournalVolNumPages{Applied Energy}{333}{}{120565}.
\newblock
\begin{APACrefURL} [{2024-09-24}]\url{https://www.sciencedirect.com/science/article/pii/S0306261922018220} \end{APACrefURL}
\newblock
\begin{APACrefDOI} \doi{10.1016/j.apenergy.2022.120565} \end{APACrefDOI}
\PrintBackRefs{\CurrentBib}

\bibitem [\protect \citeauthoryear {%
Bi%
\ \protect \BOthers {.}}{%
Bi%
\ \protect \BOthers {.}}{%
{\protect \APACyear {2023}}%
}]{%
bi2023}
\APACinsertmetastar {%
bi2023}%
\begin{APACrefauthors}%
Bi, K.%
, Xie, L.%
, Zhang, H.%
, Chen, X.%
, Gu, X.%
\BCBL {}\ \BBA {} Tian, Q.%
\end{APACrefauthors}%
\unskip\
\newblock
\APACrefYearMonthDay{2023}{{\APACmonth{07}}}{}.
\newblock
{\BBOQ}\APACrefatitle {Accurate medium-range global weather forecasting with {3D} neural networks} {Accurate medium-range global weather forecasting with {3D} neural networks}.{\BBCQ}
\newblock
\APACjournalVolNumPages{Nature}{619}{7970}{533--538}.
\newblock
\begin{APACrefURL} [{2024-09-24}]\url{https://www.nature.com/articles/s41586-023-06185-3} \end{APACrefURL}
\newblock
\APACrefnote{Publisher: Nature Publishing Group}
\newblock
\begin{APACrefDOI} \doi{10.1038/s41586-023-06185-3} \end{APACrefDOI}
\PrintBackRefs{\CurrentBib}

\bibitem [\protect \citeauthoryear {%
Bodnar%
\ \protect \BOthers {.}}{%
Bodnar%
\ \protect \BOthers {.}}{%
{\protect \APACyear {2024}}%
}]{%
bodnar2024}
\APACinsertmetastar {%
bodnar2024}%
\begin{APACrefauthors}%
Bodnar, C.%
, Bruinsma, W\BPBI P.%
, Lucic, A.%
, Stanley, M.%
, Brandstetter, J.%
, Garvan, P.%
\BDBL {}Perdikaris, P.%
\end{APACrefauthors}%
\unskip\
\newblock
\APACrefYearMonthDay{2024}{}{}.
\newblock
\APACrefbtitle {Aurora: {A} {Foundation} {Model} of the {Atmosphere}.} {Aurora: {A} {Foundation} {Model} of the {Atmosphere}.}
\PrintBackRefs{\CurrentBib}

\bibitem [\protect \citeauthoryear {%
Boussif%
, Assouline%
, Benabbou%
\BCBL {}\ \BBA {} Bengio%
}{%
Boussif%
\ \protect \BOthers {.}}{%
{\protect \APACyear {2022}}%
}]{%
boussif2022}
\APACinsertmetastar {%
boussif2022}%
\begin{APACrefauthors}%
Boussif, O.%
, Assouline, D.%
, Benabbou, L.%
\BCBL {}\ \BBA {} Bengio, Y.%
\end{APACrefauthors}%
\unskip\
\newblock
\APACrefYearMonthDay{2022}{{\APACmonth{10}}}{}.
\newblock
\APACrefbtitle {{MAgNet}: {Mesh} {Agnostic} {Neural} {PDE} {Solver}.} {{MAgNet}: {Mesh} {Agnostic} {Neural} {PDE} {Solver}.}
\newblock
\APACaddressPublisher{}{arXiv}.
\newblock
\begin{APACrefURL} [{2023-11-28}]\url{http://arxiv.org/abs/2210.05495} \end{APACrefURL}
\newblock
\APACrefnote{arXiv:2210.05495 [physics]}
\PrintBackRefs{\CurrentBib}

\bibitem [\protect \citeauthoryear {%
Brandstetter%
, Worrall%
\BCBL {}\ \BBA {} Welling%
}{%
Brandstetter%
\ \protect \BOthers {.}}{%
{\protect \APACyear {2023}}%
}]{%
brandstetter2023}
\APACinsertmetastar {%
brandstetter2023}%
\begin{APACrefauthors}%
Brandstetter, J.%
, Worrall, D.%
\BCBL {}\ \BBA {} Welling, M.%
\end{APACrefauthors}%
\unskip\
\newblock
\APACrefYearMonthDay{2023}{{\APACmonth{03}}}{}.
\newblock
\APACrefbtitle {Message {Passing} {Neural} {PDE} {Solvers}.} {Message {Passing} {Neural} {PDE} {Solvers}.}
\newblock
\APACaddressPublisher{}{arXiv}.
\newblock
\begin{APACrefURL} [{2024-05-07}]\url{http://arxiv.org/abs/2202.03376} \end{APACrefURL}
\newblock
\APACrefnote{arXiv:2202.03376 [cs, math]}
\PrintBackRefs{\CurrentBib}

\bibitem [\protect \citeauthoryear {%
Delaunay%
}{%
Delaunay%
}{%
{\protect \APACyear {1934}}%
}]{%
delaunay1934}
\APACinsertmetastar {%
delaunay1934}%
\begin{APACrefauthors}%
Delaunay, B\BPBI N.%
\end{APACrefauthors}%
\unskip\
\newblock
\APACrefYearMonthDay{1934}{}{}.
\newblock
{\BBOQ}\APACrefatitle {Sur la sphère vide. {A} la mémoire de {Georges} {Voronoï}} {Sur la sphère vide. {A} la mémoire de {Georges} {Voronoï}}.{\BBCQ}
\newblock
\APACjournalVolNumPages{Известия Российской академии наук. Серия математическая}{6}{}{793--800}.
\PrintBackRefs{\CurrentBib}

\bibitem [\protect \citeauthoryear {%
Dosovitskiy%
\ \protect \BOthers {.}}{%
Dosovitskiy%
\ \protect \BOthers {.}}{%
{\protect \APACyear {2021}}%
}]{%
dosovitskiy2021a}
\APACinsertmetastar {%
dosovitskiy2021a}%
\begin{APACrefauthors}%
Dosovitskiy, A.%
, Beyer, L.%
, Kolesnikov, A.%
, Weissenborn, D.%
, Zhai, X.%
, Unterthiner, T.%
\BDBL {}Houlsby, N.%
\end{APACrefauthors}%
\unskip\
\newblock
\APACrefYearMonthDay{2021}{{\APACmonth{06}}}{}.
\newblock
\APACrefbtitle {An {Image} is {Worth} 16x16 {Words}: {Transformers} for {Image} {Recognition} at {Scale}.} {An {Image} is {Worth} 16x16 {Words}: {Transformers} for {Image} {Recognition} at {Scale}.}
\newblock
\APACaddressPublisher{}{arXiv}.
\newblock
\begin{APACrefURL} [{2023-11-03}]\url{http://arxiv.org/abs/2010.11929} \end{APACrefURL}
\newblock
\APACrefnote{arXiv:2010.11929 [cs]}
\PrintBackRefs{\CurrentBib}

\bibitem [\protect \citeauthoryear {%
Dowell%
\ \protect \BOthers {.}}{%
Dowell%
\ \protect \BOthers {.}}{%
{\protect \APACyear {2022}}%
}]{%
dowell2022}
\APACinsertmetastar {%
dowell2022}%
\begin{APACrefauthors}%
Dowell, D\BPBI C.%
, Alexander, C\BPBI R.%
, James, E\BPBI P.%
, Weygandt, S\BPBI S.%
, Benjamin, S\BPBI G.%
, Manikin, G\BPBI S.%
\BDBL {}Alcott, T\BPBI I.%
\end{APACrefauthors}%
\unskip\
\newblock
\APACrefYearMonthDay{2022}{{\APACmonth{08}}}{}.
\newblock
{\BBOQ}\APACrefatitle {The {High}-{Resolution} {Rapid} {Refresh} ({HRRR}): {An} {Hourly} {Updating} {Convection}-{Allowing} {Forecast} {Model}. {Part} {I}: {Motivation} and {System} {Description}} {The {High}-{Resolution} {Rapid} {Refresh} ({HRRR}): {An} {Hourly} {Updating} {Convection}-{Allowing} {Forecast} {Model}. {Part} {I}: {Motivation} and {System} {Description}}.{\BBCQ}
\newblock
\APACjournalVolNumPages{Weather and Forecasting}{37}{8}{1371--1395}.
\newblock
\begin{APACrefURL} [{2024-09-28}]\url{https://journals.ametsoc.org/view/journals/wefo/37/8/WAF-D-21-0151.1.xml} \end{APACrefURL}
\newblock
\begin{APACrefDOI} \doi{10.1175/WAF-D-21-0151.1} \end{APACrefDOI}
\PrintBackRefs{\CurrentBib}

\bibitem [\protect \citeauthoryear {%
ECMWF%
}{%
ECMWF%
}{%
{\protect \APACyear {2022}}%
}]{%
ECMWF2022}
\APACinsertmetastar {%
ECMWF2022}%
\begin{APACrefauthors}%
ECMWF.%
\end{APACrefauthors}%
\unskip\
\newblock
\APACrefYearMonthDay{2022}{{\APACmonth{05}}}{}.
\newblock
\APACrefbtitle {Integrated {Forecasting} {System}} {Integrated {Forecasting} {System}}\ [Text].
\newblock
\begin{APACrefURL} [{2024-09-28}]\url{https://www.ecmwf.int/en/forecasts/documentation-and-support/changes-ecmwf-model} \end{APACrefURL}
\PrintBackRefs{\CurrentBib}

\bibitem [\protect \citeauthoryear {%
Flora%
\ \BBA {} Potvin%
}{%
Flora%
\ \BBA {} Potvin%
}{%
{\protect \APACyear {2025}}%
}]{%
flora2025}
\APACinsertmetastar {%
flora2025}%
\begin{APACrefauthors}%
Flora, M\BPBI L.%
\BCBT {}\ \BBA {} Potvin, C.%
\end{APACrefauthors}%
\unskip\
\newblock
\APACrefYearMonthDay{2025}{}{}.
\newblock
{\BBOQ}\APACrefatitle {{WoFSCast}: {A} {Machine} {Learning} {Model} for {Predicting} {Thunderstorms} at {Watch}-to-{Warning} {Scales}} {{WoFSCast}: {A} {Machine} {Learning} {Model} for {Predicting} {Thunderstorms} at {Watch}-to-{Warning} {Scales}}.{\BBCQ}
\newblock
\APACjournalVolNumPages{Geophysical Research Letters}{52}{10}{e2024GL112383}.
\newblock
\begin{APACrefURL} [{2025-07-28}]\url{https://onlinelibrary.wiley.com/doi/abs/10.1029/2024GL112383} \end{APACrefURL}
\newblock
\APACrefnote{\_eprint: https://agupubs.onlinelibrary.wiley.com/doi/pdf/10.1029/2024GL112383}
\newblock
\begin{APACrefDOI} \doi{10.1029/2024GL112383} \end{APACrefDOI}
\PrintBackRefs{\CurrentBib}

\bibitem [\protect \citeauthoryear {%
Gilmer%
, Schoenholz%
, Riley%
, Vinyals%
\BCBL {}\ \BBA {} Dahl%
}{%
Gilmer%
\ \protect \BOthers {.}}{%
{\protect \APACyear {2017}}%
}]{%
gilmer2017}
\APACinsertmetastar {%
gilmer2017}%
\begin{APACrefauthors}%
Gilmer, J.%
, Schoenholz, S\BPBI S.%
, Riley, P\BPBI F.%
, Vinyals, O.%
\BCBL {}\ \BBA {} Dahl, G\BPBI E.%
\end{APACrefauthors}%
\unskip\
\newblock
\APACrefYearMonthDay{2017}{{\APACmonth{06}}}{}.
\newblock
\APACrefbtitle {Neural {Message} {Passing} for {Quantum} {Chemistry}.} {Neural {Message} {Passing} for {Quantum} {Chemistry}.}
\newblock
\APACaddressPublisher{}{arXiv}.
\newblock
\begin{APACrefURL} [{2024-09-25}]\url{http://arxiv.org/abs/1704.01212} \end{APACrefURL}
\newblock
\APACrefnote{arXiv:1704.01212 [cs]}
\newblock
\begin{APACrefDOI} \doi{10.48550/arXiv.1704.01212} \end{APACrefDOI}
\PrintBackRefs{\CurrentBib}

\bibitem [\protect \citeauthoryear {%
Giorgi%
\ \protect \BOthers {.}}{%
Giorgi%
\ \protect \BOthers {.}}{%
{\protect \APACyear {2012}}%
}]{%
giorgi2012}
\APACinsertmetastar {%
giorgi2012}%
\begin{APACrefauthors}%
Giorgi, F.%
, Coppola, E.%
, Solmon, F.%
, Mariotti, L.%
, Sylla, M\BPBI B.%
, Bi, X.%
\BDBL {}Brankovic, C.%
\end{APACrefauthors}%
\unskip\
\newblock
\APACrefYearMonthDay{2012}{{\APACmonth{03}}}{}.
\newblock
{\BBOQ}\APACrefatitle {{RegCM4}: model description and preliminary tests over multiple {CORDEX} domains} {{RegCM4}: model description and preliminary tests over multiple {CORDEX} domains}.{\BBCQ}
\newblock
\APACjournalVolNumPages{Climate Research}{52}{}{7--29}.
\newblock
\begin{APACrefURL} [{2025-03-06}]\url{https://www.int-res.com/abstracts/cr/v52/p7-29/} \end{APACrefURL}
\newblock
\begin{APACrefDOI} \doi{10.3354/cr01018} \end{APACrefDOI}
\PrintBackRefs{\CurrentBib}

\bibitem [\protect \citeauthoryear {%
Harder%
\ \protect \BOthers {.}}{%
Harder%
\ \protect \BOthers {.}}{%
{\protect \APACyear {2023}}%
}]{%
harder2023}
\APACinsertmetastar {%
harder2023}%
\begin{APACrefauthors}%
Harder, P.%
, Hernandez-Garcia, A.%
, Ramesh, V.%
, Yang, Q.%
, Sattegeri, P.%
, Szwarcman, D.%
\BDBL {}Rolnick, D.%
\end{APACrefauthors}%
\unskip\
\newblock
\APACrefYearMonthDay{2023}{}{}.
\newblock
{\BBOQ}\APACrefatitle {Hard-{Constrained} {Deep} {Learning} for {Climate} {Downscaling}} {Hard-{Constrained} {Deep} {Learning} for {Climate} {Downscaling}}.{\BBCQ}
\newblock
\APACjournalVolNumPages{Journal of Machine Learning Research}{24}{365}{1--40}.
\newblock
\begin{APACrefURL} [{2024-10-01}]\url{http://jmlr.org/papers/v24/23-0158.html} \end{APACrefURL}
\PrintBackRefs{\CurrentBib}

\bibitem [\protect \citeauthoryear {%
Hernanz%
\ \protect \BOthers {.}}{%
Hernanz%
\ \protect \BOthers {.}}{%
{\protect \APACyear {2022}}%
}]{%
hernanz2022}
\APACinsertmetastar {%
hernanz2022}%
\begin{APACrefauthors}%
Hernanz, A.%
, García-Valero, J\BPBI A.%
, Domínguez, M.%
, Ramos-Calzado, P.%
, Pastor-Saavedra, M\BPBI A.%
\BCBL {}\ \BBA {} Rodríguez-Camino, E.%
\end{APACrefauthors}%
\unskip\
\newblock
\APACrefYearMonthDay{2022}{}{}.
\newblock
{\BBOQ}\APACrefatitle {Evaluation of statistical downscaling methods for climate change projections over {Spain}: {Present} conditions with perfect predictors} {Evaluation of statistical downscaling methods for climate change projections over {Spain}: {Present} conditions with perfect predictors}.{\BBCQ}
\newblock
\APACjournalVolNumPages{International Journal of Climatology}{42}{2}{762--776}.
\newblock
\begin{APACrefURL} [{2025-03-06}]\url{https://onlinelibrary.wiley.com/doi/abs/10.1002/joc.7271} \end{APACrefURL}
\newblock
\APACrefnote{\_eprint: https://onlinelibrary.wiley.com/doi/pdf/10.1002/joc.7271}
\newblock
\begin{APACrefDOI} \doi{10.1002/joc.7271} \end{APACrefDOI}
\PrintBackRefs{\CurrentBib}

\bibitem [\protect \citeauthoryear {%
Hersbach%
\ \protect \BOthers {.}}{%
Hersbach%
\ \protect \BOthers {.}}{%
{\protect \APACyear {2020}}%
}]{%
hersbach2020}
\APACinsertmetastar {%
hersbach2020}%
\begin{APACrefauthors}%
Hersbach, H.%
, Bell, B.%
, Berrisford, P.%
, Hirahara, S.%
, Horányi, A.%
, Muñoz-Sabater, J.%
\BDBL {}Thépaut, J\BHBI N.%
\end{APACrefauthors}%
\unskip\
\newblock
\APACrefYearMonthDay{2020}{}{}.
\newblock
{\BBOQ}\APACrefatitle {The {ERA5} global reanalysis} {The {ERA5} global reanalysis}.{\BBCQ}
\newblock
\APACjournalVolNumPages{Quarterly Journal of the Royal Meteorological Society}{146}{730}{1999--2049}.
\newblock
\begin{APACrefURL} [{2024-09-23}]\url{https://onlinelibrary.wiley.com/doi/abs/10.1002/qj.3803} \end{APACrefURL}
\newblock
\APACrefnote{\_eprint: https://onlinelibrary.wiley.com/doi/pdf/10.1002/qj.3803}
\newblock
\begin{APACrefDOI} \doi{10.1002/qj.3803} \end{APACrefDOI}
\PrintBackRefs{\CurrentBib}

\bibitem [\protect \citeauthoryear {%
Kingma%
\ \BBA {} Ba%
}{%
Kingma%
\ \BBA {} Ba%
}{%
{\protect \APACyear {2017}}%
}]{%
kingma2017}
\APACinsertmetastar {%
kingma2017}%
\begin{APACrefauthors}%
Kingma, D\BPBI P.%
\BCBT {}\ \BBA {} Ba, J.%
\end{APACrefauthors}%
\unskip\
\newblock
\APACrefYearMonthDay{2017}{{\APACmonth{01}}}{}.
\newblock
\APACrefbtitle {Adam: {A} {Method} for {Stochastic} {Optimization}.} {Adam: {A} {Method} for {Stochastic} {Optimization}.}
\newblock
\APACaddressPublisher{}{arXiv}.
\newblock
\begin{APACrefURL} [{2025-08-17}]\url{http://arxiv.org/abs/1412.6980} \end{APACrefURL}
\newblock
\APACrefnote{arXiv:1412.6980 [cs]}
\newblock
\begin{APACrefDOI} \doi{10.48550/arXiv.1412.6980} \end{APACrefDOI}
\PrintBackRefs{\CurrentBib}

\bibitem [\protect \citeauthoryear {%
Kochkov%
\ \protect \BOthers {.}}{%
Kochkov%
\ \protect \BOthers {.}}{%
{\protect \APACyear {2024}}%
}]{%
kochkov2024}
\APACinsertmetastar {%
kochkov2024}%
\begin{APACrefauthors}%
Kochkov, D.%
, Yuval, J.%
, Langmore, I.%
, Norgaard, P.%
, Smith, J.%
, Mooers, G.%
\BDBL {}Hoyer, S.%
\end{APACrefauthors}%
\unskip\
\newblock
\APACrefYearMonthDay{2024}{{\APACmonth{07}}}{}.
\newblock
{\BBOQ}\APACrefatitle {Neural general circulation models for weather and climate} {Neural general circulation models for weather and climate}.{\BBCQ}
\newblock
\APACjournalVolNumPages{Nature}{}{}{}.
\newblock
\begin{APACrefURL} [{2024-07-23}]\url{https://www.nature.com/articles/s41586-024-07744-y} \end{APACrefURL}
\newblock
\begin{APACrefDOI} \doi{10.1038/s41586-024-07744-y} \end{APACrefDOI}
\PrintBackRefs{\CurrentBib}

\bibitem [\protect \citeauthoryear {%
Lam%
\ \protect \BOthers {.}}{%
Lam%
\ \protect \BOthers {.}}{%
{\protect \APACyear {2023}}%
}]{%
lam2023}
\APACinsertmetastar {%
lam2023}%
\begin{APACrefauthors}%
Lam, R.%
, Sanchez-Gonzalez, A.%
, Willson, M.%
, Wirnsberger, P.%
, Fortunato, M.%
, Alet, F.%
\BDBL {}Battaglia, P.%
\end{APACrefauthors}%
\unskip\
\newblock
\APACrefYearMonthDay{2023}{{\APACmonth{12}}}{}.
\newblock
{\BBOQ}\APACrefatitle {Learning skillful medium-range global weather forecasting} {Learning skillful medium-range global weather forecasting}.{\BBCQ}
\newblock
\APACjournalVolNumPages{Science}{382}{6677}{1416--1421}.
\newblock
\begin{APACrefURL} [{2024-03-09}]\url{https://www.science.org/doi/10.1126/science.adi2336} \end{APACrefURL}
\newblock
\begin{APACrefDOI} \doi{10.1126/science.adi2336} \end{APACrefDOI}
\PrintBackRefs{\CurrentBib}

\bibitem [\protect \citeauthoryear {%
Lang%
\ \protect \BOthers {.}}{%
Lang%
\ \protect \BOthers {.}}{%
{\protect \APACyear {2024}}%
}]{%
lang2024}
\APACinsertmetastar {%
lang2024}%
\begin{APACrefauthors}%
Lang, S.%
, Alexe, M.%
, Chantry, M.%
, Dramsch, J.%
, Pinault, F.%
, Raoult, B.%
\BDBL {}Rabier, F.%
\end{APACrefauthors}%
\unskip\
\newblock
\APACrefYearMonthDay{2024}{{\APACmonth{06}}}{}.
\newblock
\APACrefbtitle {{AIFS} - {ECMWF}'s data-driven forecasting system.} {{AIFS} - {ECMWF}'s data-driven forecasting system.}
\newblock
\APACaddressPublisher{}{arXiv}.
\newblock
\begin{APACrefURL} [{2024-06-19}]\url{http://arxiv.org/abs/2406.01465} \end{APACrefURL}
\newblock
\APACrefnote{arXiv:2406.01465 [physics]}
\PrintBackRefs{\CurrentBib}

\bibitem [\protect \citeauthoryear {%
Lessig%
\ \protect \BOthers {.}}{%
Lessig%
\ \protect \BOthers {.}}{%
{\protect \APACyear {2023}}%
}]{%
lessig2023}
\APACinsertmetastar {%
lessig2023}%
\begin{APACrefauthors}%
Lessig, C.%
, Luise, I.%
, Gong, B.%
, Langguth, M.%
, Stadtler, S.%
\BCBL {}\ \BBA {} Schultz, M.%
\end{APACrefauthors}%
\unskip\
\newblock
\APACrefYearMonthDay{2023}{{\APACmonth{09}}}{}.
\newblock
\APACrefbtitle {{AtmoRep}: {A} stochastic model of atmosphere dynamics using large scale representation learning.} {{AtmoRep}: {A} stochastic model of atmosphere dynamics using large scale representation learning.}
\newblock
\APACaddressPublisher{}{arXiv}.
\newblock
\begin{APACrefURL} [{2024-03-09}]\url{http://arxiv.org/abs/2308.13280} \end{APACrefURL}
\newblock
\APACrefnote{arXiv:2308.13280 [physics]}
\PrintBackRefs{\CurrentBib}

\bibitem [\protect \citeauthoryear {%
Mouatadid%
\ \protect \BOthers {.}}{%
Mouatadid%
\ \protect \BOthers {.}}{%
{\protect \APACyear {2023}}%
}]{%
mouatadid2023}
\APACinsertmetastar {%
mouatadid2023}%
\begin{APACrefauthors}%
Mouatadid, S.%
, Orenstein, P.%
, Flaspohler, G.%
, Cohen, J.%
, Oprescu, M.%
, Fraenkel, E.%
\BCBL {}\ \BBA {} Mackey, L.%
\end{APACrefauthors}%
\unskip\
\newblock
\APACrefYearMonthDay{2023}{{\APACmonth{06}}}{}.
\newblock
{\BBOQ}\APACrefatitle {Adaptive bias correction for improved subseasonal forecasting} {Adaptive bias correction for improved subseasonal forecasting}.{\BBCQ}
\newblock
\APACjournalVolNumPages{Nature Communications}{14}{1}{3482}.
\newblock
\begin{APACrefURL} [{2024-11-18}]\url{https://www.nature.com/articles/s41467-023-38874-y} \end{APACrefURL}
\newblock
\begin{APACrefDOI} \doi{10.1038/s41467-023-38874-y} \end{APACrefDOI}
\PrintBackRefs{\CurrentBib}

\bibitem [\protect \citeauthoryear {%
Nguyen%
, Brandstetter%
, Kapoor%
, Gupta%
\BCBL {}\ \BBA {} Grover%
}{%
Nguyen%
\ \protect \BOthers {.}}{%
{\protect \APACyear {2023}}%
}]{%
nguyen2023}
\APACinsertmetastar {%
nguyen2023}%
\begin{APACrefauthors}%
Nguyen, T.%
, Brandstetter, J.%
, Kapoor, A.%
, Gupta, J\BPBI K.%
\BCBL {}\ \BBA {} Grover, A.%
\end{APACrefauthors}%
\unskip\
\newblock
\APACrefYearMonthDay{2023}{{\APACmonth{12}}}{}.
\newblock
\APACrefbtitle {{ClimaX}: {A} foundation model for weather and climate.} {{ClimaX}: {A} foundation model for weather and climate.}
\newblock
\APACaddressPublisher{}{arXiv}.
\newblock
\begin{APACrefURL} [{2024-03-09}]\url{http://arxiv.org/abs/2301.10343} \end{APACrefURL}
\newblock
\APACrefnote{arXiv:2301.10343 [cs]}
\PrintBackRefs{\CurrentBib}

\bibitem [\protect \citeauthoryear {%
Oskarsson%
, Landelius%
\BCBL {}\ \BBA {} Lindsten%
}{%
Oskarsson%
\ \protect \BOthers {.}}{%
{\protect \APACyear {2023}}%
}]{%
oskarsson2023}
\APACinsertmetastar {%
oskarsson2023}%
\begin{APACrefauthors}%
Oskarsson, J.%
, Landelius, T.%
\BCBL {}\ \BBA {} Lindsten, F.%
\end{APACrefauthors}%
\unskip\
\newblock
\APACrefYearMonthDay{2023}{{\APACmonth{11}}}{}.
\newblock
\APACrefbtitle {Graph-based {Neural} {Weather} {Prediction} for {Limited} {Area} {Modeling}.} {Graph-based {Neural} {Weather} {Prediction} for {Limited} {Area} {Modeling}.}
\newblock
\APACaddressPublisher{}{arXiv}.
\newblock
\begin{APACrefURL} [{2024-06-03}]\url{http://arxiv.org/abs/2309.17370} \end{APACrefURL}
\newblock
\APACrefnote{arXiv:2309.17370 [cs, stat]}
\PrintBackRefs{\CurrentBib}

\bibitem [\protect \citeauthoryear {%
Pathak%
\ \protect \BOthers {.}}{%
Pathak%
\ \protect \BOthers {.}}{%
{\protect \APACyear {2024}}%
}]{%
pathak2024}
\APACinsertmetastar {%
pathak2024}%
\begin{APACrefauthors}%
Pathak, J.%
, Cohen, Y.%
, Garg, P.%
, Harrington, P.%
, Brenowitz, N.%
, Durran, D.%
\BDBL {}Pritchard, M.%
\end{APACrefauthors}%
\unskip\
\newblock
\APACrefYearMonthDay{2024}{{\APACmonth{08}}}{}.
\newblock
\APACrefbtitle {Kilometer-{Scale} {Convection} {Allowing} {Model} {Emulation} using {Generative} {Diffusion} {Modeling}.} {Kilometer-{Scale} {Convection} {Allowing} {Model} {Emulation} using {Generative} {Diffusion} {Modeling}.}
\newblock
\APACaddressPublisher{}{arXiv}.
\newblock
\begin{APACrefURL} [{2025-07-28}]\url{http://arxiv.org/abs/2408.10958} \end{APACrefURL}
\newblock
\APACrefnote{arXiv:2408.10958 [physics]}
\newblock
\begin{APACrefDOI} \doi{10.48550/arXiv.2408.10958} \end{APACrefDOI}
\PrintBackRefs{\CurrentBib}

\bibitem [\protect \citeauthoryear {%
Pathak%
\ \protect \BOthers {.}}{%
Pathak%
\ \protect \BOthers {.}}{%
{\protect \APACyear {2022}}%
}]{%
pathak2022}
\APACinsertmetastar {%
pathak2022}%
\begin{APACrefauthors}%
Pathak, J.%
, Subramanian, S.%
, Harrington, P.%
, Raja, S.%
, Chattopadhyay, A.%
, Mardani, M.%
\BDBL {}Anandkumar, A.%
\end{APACrefauthors}%
\unskip\
\newblock
\APACrefYearMonthDay{2022}{{\APACmonth{02}}}{}.
\newblock
\APACrefbtitle {{FourCastNet}: {A} {Global} {Data}-driven {High}-resolution {Weather} {Model} using {Adaptive} {Fourier} {Neural} {Operators}.} {{FourCastNet}: {A} {Global} {Data}-driven {High}-resolution {Weather} {Model} using {Adaptive} {Fourier} {Neural} {Operators}.}
\newblock
\APACaddressPublisher{}{arXiv}.
\newblock
\begin{APACrefURL} [{2024-04-15}]\url{http://arxiv.org/abs/2202.11214} \end{APACrefURL}
\newblock
\APACrefnote{arXiv:2202.11214 [physics]}
\PrintBackRefs{\CurrentBib}

\bibitem [\protect \citeauthoryear {%
Pfaff%
, Fortunato%
, Sanchez-Gonzalez%
\BCBL {}\ \BBA {} Battaglia%
}{%
Pfaff%
\ \protect \BOthers {.}}{%
{\protect \APACyear {2021}}%
}]{%
pfaff2021}
\APACinsertmetastar {%
pfaff2021}%
\begin{APACrefauthors}%
Pfaff, T.%
, Fortunato, M.%
, Sanchez-Gonzalez, A.%
\BCBL {}\ \BBA {} Battaglia, P\BPBI W.%
\end{APACrefauthors}%
\unskip\
\newblock
\APACrefYearMonthDay{2021}{{\APACmonth{06}}}{}.
\newblock
\APACrefbtitle {Learning {Mesh}-{Based} {Simulation} with {Graph} {Networks}.} {Learning {Mesh}-{Based} {Simulation} with {Graph} {Networks}.}
\newblock
\APACaddressPublisher{}{arXiv}.
\newblock
\begin{APACrefURL} [{2024-07-23}]\url{http://arxiv.org/abs/2010.03409} \end{APACrefURL}
\newblock
\APACrefnote{arXiv:2010.03409 [cs]}
\PrintBackRefs{\CurrentBib}

\bibitem [\protect \citeauthoryear {%
Prasad%
\ \protect \BOthers {.}}{%
Prasad%
\ \protect \BOthers {.}}{%
{\protect \APACyear {2024}}%
}]{%
prasad2024}
\APACinsertmetastar {%
prasad2024}%
\begin{APACrefauthors}%
Prasad, A.%
, Harder, P.%
, Yang, Q.%
, Sattegeri, P.%
, Szwarcman, D.%
, Watson, C.%
\BCBL {}\ \BBA {} Rolnick, D.%
\end{APACrefauthors}%
\unskip\
\newblock
\APACrefYearMonthDay{2024}{{\APACmonth{07}}}{}.
\newblock
\APACrefbtitle {Evaluating the transferability potential of deep learning models for climate downscaling.} {Evaluating the transferability potential of deep learning models for climate downscaling.}
\newblock
\APACaddressPublisher{}{arXiv}.
\newblock
\begin{APACrefURL} [{2024-10-01}]\url{http://arxiv.org/abs/2407.12517} \end{APACrefURL}
\newblock
\APACrefnote{arXiv:2407.12517 [cs]}
\newblock
\begin{APACrefDOI} \doi{10.48550/arXiv.2407.12517} \end{APACrefDOI}
\PrintBackRefs{\CurrentBib}

\bibitem [\protect \citeauthoryear {%
Ramavajjala%
\ \BBA {} Mitra%
}{%
Ramavajjala%
\ \BBA {} Mitra%
}{%
{\protect \APACyear {2023}}%
}]{%
ramavajjala2023}
\APACinsertmetastar {%
ramavajjala2023}%
\begin{APACrefauthors}%
Ramavajjala, V.%
\BCBT {}\ \BBA {} Mitra, P\BPBI P.%
\end{APACrefauthors}%
\unskip\
\newblock
\APACrefYearMonthDay{2023}{{\APACmonth{09}}}{}.
\newblock
\APACrefbtitle {Verification against in-situ observations for {Data}-{Driven} {Weather} {Prediction}.} {Verification against in-situ observations for {Data}-{Driven} {Weather} {Prediction}.}
\newblock
\APACaddressPublisher{}{arXiv}.
\newblock
\begin{APACrefURL} [{2024-01-22}]\url{http://arxiv.org/abs/2305.00048} \end{APACrefURL}
\newblock
\APACrefnote{arXiv:2305.00048 [physics]}
\newblock
\begin{APACrefDOI} \doi{10.48550/arXiv.2305.00048} \end{APACrefDOI}
\PrintBackRefs{\CurrentBib}

\bibitem [\protect \citeauthoryear {%
Ronneberger%
, Fischer%
\BCBL {}\ \BBA {} Brox%
}{%
Ronneberger%
\ \protect \BOthers {.}}{%
{\protect \APACyear {2015}}%
}]{%
ronneberger2015}
\APACinsertmetastar {%
ronneberger2015}%
\begin{APACrefauthors}%
Ronneberger, O.%
, Fischer, P.%
\BCBL {}\ \BBA {} Brox, T.%
\end{APACrefauthors}%
\unskip\
\newblock
\APACrefYearMonthDay{2015}{{\APACmonth{05}}}{}.
\newblock
\APACrefbtitle {U-{Net}: {Convolutional} {Networks} for {Biomedical} {Image} {Segmentation}.} {U-{Net}: {Convolutional} {Networks} for {Biomedical} {Image} {Segmentation}.}
\newblock
\APACaddressPublisher{}{arXiv}.
\newblock
\begin{APACrefURL} [{2024-09-24}]\url{http://arxiv.org/abs/1505.04597} \end{APACrefURL}
\newblock
\APACrefnote{arXiv:1505.04597 [cs]}
\newblock
\begin{APACrefDOI} \doi{10.48550/arXiv.1505.04597} \end{APACrefDOI}
\PrintBackRefs{\CurrentBib}

\bibitem [\protect \citeauthoryear {%
Sanchez-Gonzalez%
\ \protect \BOthers {.}}{%
Sanchez-Gonzalez%
\ \protect \BOthers {.}}{%
{\protect \APACyear {2020}}%
}]{%
sanchez-gonzalez2020}
\APACinsertmetastar {%
sanchez-gonzalez2020}%
\begin{APACrefauthors}%
Sanchez-Gonzalez, A.%
, Godwin, J.%
, Pfaff, T.%
, Ying, R.%
, Leskovec, J.%
\BCBL {}\ \BBA {} Battaglia, P\BPBI W.%
\end{APACrefauthors}%
\unskip\
\newblock
\APACrefYearMonthDay{2020}{{\APACmonth{09}}}{}.
\newblock
\APACrefbtitle {Learning to {Simulate} {Complex} {Physics} with {Graph} {Networks}.} {Learning to {Simulate} {Complex} {Physics} with {Graph} {Networks}.}
\newblock
\APACaddressPublisher{}{arXiv}.
\newblock
\begin{APACrefURL} [{2024-09-24}]\url{http://arxiv.org/abs/2002.09405} \end{APACrefURL}
\newblock
\APACrefnote{arXiv:2002.09405 [physics, stat]}
\newblock
\begin{APACrefDOI} \doi{10.48550/arXiv.2002.09405} \end{APACrefDOI}
\PrintBackRefs{\CurrentBib}

\bibitem [\protect \citeauthoryear {%
Schmude%
\ \protect \BOthers {.}}{%
Schmude%
\ \protect \BOthers {.}}{%
{\protect \APACyear {2024}}%
}]{%
schmude2024}
\APACinsertmetastar {%
schmude2024}%
\begin{APACrefauthors}%
Schmude, J.%
, Roy, S.%
, Trojak, W.%
, Jakubik, J.%
, Civitarese, D\BPBI S.%
, Singh, S.%
\BDBL {}Ramachandran, R.%
\end{APACrefauthors}%
\unskip\
\newblock
\APACrefYearMonthDay{2024}{{\APACmonth{09}}}{}.
\newblock
\APACrefbtitle {Prithvi {WxC}: {Foundation} {Model} for {Weather} and {Climate}.} {Prithvi {WxC}: {Foundation} {Model} for {Weather} and {Climate}.}
\newblock
\APACaddressPublisher{}{arXiv}.
\newblock
\begin{APACrefURL} [{2024-09-24}]\url{http://arxiv.org/abs/2409.13598} \end{APACrefURL}
\newblock
\APACrefnote{arXiv:2409.13598 [physics]}
\PrintBackRefs{\CurrentBib}

\bibitem [\protect \citeauthoryear {%
Skamarock%
\ \protect \BOthers {.}}{%
Skamarock%
\ \protect \BOthers {.}}{%
{\protect \APACyear {2008}}%
}]{%
skamarock}
\APACinsertmetastar {%
skamarock}%
\begin{APACrefauthors}%
Skamarock, W.%
, Klemp, J.%
, Dudhia, J.%
, Gill, D\BPBI O.%
, Barker, A\BPBI D.%
, Duda, M\BPBI G.%
\BDBL {}Powers, J\BPBI G.%
\end{APACrefauthors}%
\unskip\
\newblock
\APACrefYearMonthDay{2008}{}{}.
\newblock
\APACrefbtitle {A {Description} of the {Advanced} {Research} {WRF} {Version} 3} {A {Description} of the {Advanced} {Research} {WRF} {Version} 3}\ \APACbVolEdTR{}{\BTR{}}.
\newblock
\begin{APACrefURL} [{2025-03-06}]\url{https://opensky.ucar.edu/islandora/object/technotes%3A500} \end{APACrefURL}
\PrintBackRefs{\CurrentBib}

\bibitem [\protect \citeauthoryear {%
Stengel%
, Glaws%
, Hettinger%
\BCBL {}\ \BBA {} King%
}{%
Stengel%
\ \protect \BOthers {.}}{%
{\protect \APACyear {2020}}%
}]{%
stengel2020}
\APACinsertmetastar {%
stengel2020}%
\begin{APACrefauthors}%
Stengel, K.%
, Glaws, A.%
, Hettinger, D.%
\BCBL {}\ \BBA {} King, R\BPBI N.%
\end{APACrefauthors}%
\unskip\
\newblock
\APACrefYearMonthDay{2020}{{\APACmonth{07}}}{}.
\newblock
{\BBOQ}\APACrefatitle {Adversarial super-resolution of climatological wind and solar data} {Adversarial super-resolution of climatological wind and solar data}.{\BBCQ}
\newblock
\APACjournalVolNumPages{Proceedings of the National Academy of Sciences}{117}{29}{16805--16815}.
\newblock
\begin{APACrefURL} [{2025-03-06}]\url{https://www.pnas.org/doi/10.1073/pnas.1918964117} \end{APACrefURL}
\newblock
\APACrefnote{Publisher: Proceedings of the National Academy of Sciences}
\newblock
\begin{APACrefDOI} \doi{10.1073/pnas.1918964117} \end{APACrefDOI}
\PrintBackRefs{\CurrentBib}

\bibitem [\protect \citeauthoryear {%
Storch%
, Zorita%
\BCBL {}\ \BBA {} Cubasch%
}{%
Storch%
\ \protect \BOthers {.}}{%
{\protect \APACyear {1993}}%
}]{%
storch1993}
\APACinsertmetastar {%
storch1993}%
\begin{APACrefauthors}%
Storch, H\BPBI v.%
, Zorita, E.%
\BCBL {}\ \BBA {} Cubasch, U.%
\end{APACrefauthors}%
\unskip\
\newblock
\APACrefYearMonthDay{1993}{{\APACmonth{06}}}{}.
\newblock
{\BBOQ}\APACrefatitle {Downscaling of {Global} {Climate} {Change} {Estimates} to {Regional} {Scales}: {An} {Application} to {Iberian} {Rainfall} in {Wintertime}} {Downscaling of {Global} {Climate} {Change} {Estimates} to {Regional} {Scales}: {An} {Application} to {Iberian} {Rainfall} in {Wintertime}}.{\BBCQ}
\newblock
\APACjournalVolNumPages{Journal of Climate}{}{}{}.
\newblock
\begin{APACrefURL} [{2025-03-06}]\url{https://journals.ametsoc.org/view/journals/clim/6/6/1520-0442_1993_006_1161_dogcce_2_0_co_2.xml} \end{APACrefURL}
\newblock
\APACrefnote{Section: Journal of Climate}
\PrintBackRefs{\CurrentBib}

\bibitem [\protect \citeauthoryear {%
Vandal%
\ \protect \BOthers {.}}{%
Vandal%
\ \protect \BOthers {.}}{%
{\protect \APACyear {2017}}%
}]{%
vandal2017}
\APACinsertmetastar {%
vandal2017}%
\begin{APACrefauthors}%
Vandal, T.%
, Kodra, E.%
, Ganguly, S.%
, Michaelis, A.%
, Nemani, R.%
\BCBL {}\ \BBA {} Ganguly, A\BPBI R.%
\end{APACrefauthors}%
\unskip\
\newblock
\APACrefYearMonthDay{2017}{{\APACmonth{03}}}{}.
\newblock
\APACrefbtitle {{DeepSD}: {Generating} {High} {Resolution} {Climate} {Change} {Projections} through {Single} {Image} {Super}-{Resolution}.} {{DeepSD}: {Generating} {High} {Resolution} {Climate} {Change} {Projections} through {Single} {Image} {Super}-{Resolution}.}
\newblock
\APACaddressPublisher{}{arXiv}.
\newblock
\begin{APACrefURL} [{2025-03-06}]\url{http://arxiv.org/abs/1703.03126} \end{APACrefURL}
\newblock
\APACrefnote{arXiv:1703.03126 [cs]}
\newblock
\begin{APACrefDOI} \doi{10.48550/arXiv.1703.03126} \end{APACrefDOI}
\PrintBackRefs{\CurrentBib}

\bibitem [\protect \citeauthoryear {%
Vaswani%
\ \protect \BOthers {.}}{%
Vaswani%
\ \protect \BOthers {.}}{%
{\protect \APACyear {2023}}%
}]{%
vaswani2023a}
\APACinsertmetastar {%
vaswani2023a}%
\begin{APACrefauthors}%
Vaswani, A.%
, Shazeer, N.%
, Parmar, N.%
, Uszkoreit, J.%
, Jones, L.%
, Gomez, A\BPBI N.%
\BDBL {}Polosukhin, I.%
\end{APACrefauthors}%
\unskip\
\newblock
\APACrefYearMonthDay{2023}{{\APACmonth{08}}}{}.
\newblock
\APACrefbtitle {Attention {Is} {All} {You} {Need}.} {Attention {Is} {All} {You} {Need}.}
\newblock
\APACaddressPublisher{}{arXiv}.
\newblock
\begin{APACrefURL} [{2023-11-03}]\url{http://arxiv.org/abs/1706.03762} \end{APACrefURL}
\newblock
\APACrefnote{arXiv:1706.03762 [cs]}
\PrintBackRefs{\CurrentBib}

\bibitem [\protect \citeauthoryear {%
Wilby%
}{%
Wilby%
}{%
{\protect \APACyear {1998}}%
}]{%
wilby1998}
\APACinsertmetastar {%
wilby1998}%
\begin{APACrefauthors}%
Wilby, R\BPBI L.%
\end{APACrefauthors}%
\unskip\
\newblock
\APACrefYearMonthDay{1998}{{\APACmonth{12}}}{}.
\newblock
{\BBOQ}\APACrefatitle {Statistical downscaling of daily precipitation using daily airflow and seasonal teleconnection indices} {Statistical downscaling of daily precipitation using daily airflow and seasonal teleconnection indices}.{\BBCQ}
\newblock
\APACjournalVolNumPages{Climate Research}{10}{3}{163--178}.
\newblock
\begin{APACrefURL} [{2025-03-06}]\url{https://www.int-res.com/abstracts/cr/v10/n3/p163-178/} \end{APACrefURL}
\newblock
\begin{APACrefDOI} \doi{10.3354/cr010163} \end{APACrefDOI}
\PrintBackRefs{\CurrentBib}

\bibitem [\protect \citeauthoryear {%
Yang%
\ \protect \BOthers {.}}{%
Yang%
\ \protect \BOthers {.}}{%
{\protect \APACyear {2024}}%
}]{%
yang2025Dataset}
\APACinsertmetastar {%
yang2025Dataset}%
\begin{APACrefauthors}%
Yang, Q.%
, Giezendanner, J.%
, Salles~Civitarese, D.%
, Jakubik, J.%
, Schmitt, E.%
, Chandra, A.%
\BDBL {}Wang, S.%
\end{APACrefauthors}%
\unskip\
\newblock
\APACrefYearMonthDay{2024}{{\APACmonth{10}}}{}.
\newblock
\APACrefbtitle {Data for "local off-grid weather forecasting with multi-modal earth observation data".} {Data for "local off-grid weather forecasting with multi-modal earth observation data".}
\newblock
\APACaddressPublisher{}{MIT}.
\newblock
\begin{APACrefURL} \url{https://doi.org/10.5281/zenodo.16891158} \end{APACrefURL}
\newblock
\begin{APACrefDOI} \doi{10.5281/zenodo.16891158} \end{APACrefDOI}
\PrintBackRefs{\CurrentBib}

\bibitem [\protect \citeauthoryear {%
Yang%
\ \protect \BOthers {.}}{%
Yang%
\ \protect \BOthers {.}}{%
{\protect \APACyear {2025}}%
}]{%
yang2025Code}
\APACinsertmetastar {%
yang2025Code}%
\begin{APACrefauthors}%
Yang, Q.%
, Giezendanner, J.%
, Salles~Civitarese, D.%
, Jakubik, J.%
, Schmitt, E.%
, Chandra, A.%
\BDBL {}Wang, S.%
\end{APACrefauthors}%
\unskip\
\newblock
\APACrefYearMonthDay{2025}{{\APACmonth{08}}}{}.
\newblock
\APACrefbtitle {Code for "local off-grid weather forecasting with multi-modal earth observation data" ({Earth}-intelligence-lab/{LocalizedWeather}).} {Code for "local off-grid weather forecasting with multi-modal earth observation data" ({Earth}-intelligence-lab/{LocalizedWeather}).}
\newblock
\APACaddressPublisher{}{Zenodo}.
\newblock
\begin{APACrefURL} \url{https://doi.org/10.5281/zenodo.16933937} \end{APACrefURL}
\newblock
\begin{APACrefDOI} \doi{10.5281/zenodo.16933937} \end{APACrefDOI}
\PrintBackRefs{\CurrentBib}

\bibitem [\protect \citeauthoryear {%
Yang%
\ \protect \BOthers {.}}{%
Yang%
\ \protect \BOthers {.}}{%
{\protect \APACyear {2023}}%
}]{%
yang2023}
\APACinsertmetastar {%
yang2023}%
\begin{APACrefauthors}%
Yang, Q.%
, Hernandez-Garcia, A.%
, Harder, P.%
, Ramesh, V.%
, Sattegeri, P.%
, Szwarcman, D.%
\BDBL {}Rolnick, D.%
\end{APACrefauthors}%
\unskip\
\newblock
\APACrefYearMonthDay{2023}{{\APACmonth{05}}}{}.
\newblock
\APACrefbtitle {Fourier {Neural} {Operators} for {Arbitrary} {Resolution} {Climate} {Data} {Downscaling}.} {Fourier {Neural} {Operators} for {Arbitrary} {Resolution} {Climate} {Data} {Downscaling}.}
\newblock
\APACaddressPublisher{}{arXiv}.
\newblock
\begin{APACrefURL} [{2024-10-01}]\url{http://arxiv.org/abs/2305.14452} \end{APACrefURL}
\newblock
\APACrefnote{arXiv:2305.14452 [physics]}
\newblock
\begin{APACrefDOI} \doi{10.48550/arXiv.2305.14452} \end{APACrefDOI}
\PrintBackRefs{\CurrentBib}

\bibitem [\protect \citeauthoryear {%
Yang%
, Lee%
\BCBL {}\ \BBA {} Tippett%
}{%
Yang%
\ \protect \BOthers {.}}{%
{\protect \APACyear {2020}}%
}]{%
yang2020}
\APACinsertmetastar {%
yang2020}%
\begin{APACrefauthors}%
Yang, Q.%
, Lee, C\BHBI Y.%
\BCBL {}\ \BBA {} Tippett, M\BPBI K.%
\end{APACrefauthors}%
\unskip\
\newblock
\APACrefYearMonthDay{2020}{{\APACmonth{08}}}{}.
\newblock
{\BBOQ}\APACrefatitle {A {Long} {Short}-{Term} {Memory} {Model} for {Global} {Rapid} {Intensification} {Prediction}} {A {Long} {Short}-{Term} {Memory} {Model} for {Global} {Rapid} {Intensification} {Prediction}}.{\BBCQ}
\newblock
\APACjournalVolNumPages{Weather and Forecasting}{}{}{}.
\newblock
\begin{APACrefURL} [{2025-07-29}]\url{https://journals.ametsoc.org/view/journals/wefo/35/4/wafD190199.xml} \end{APACrefURL}
\newblock
\APACrefnote{Section: Weather and Forecasting}
\newblock
\begin{APACrefDOI} \doi{10.1175/WAF-D-19-0199.1} \end{APACrefDOI}
\PrintBackRefs{\CurrentBib}

\bibitem [\protect \citeauthoryear {%
Yang%
, Lee%
, Tippett%
, Chavas%
\BCBL {}\ \BBA {} Knutson%
}{%
Yang%
\ \protect \BOthers {.}}{%
{\protect \APACyear {2022}}%
}]{%
yang2022}
\APACinsertmetastar {%
yang2022}%
\begin{APACrefauthors}%
Yang, Q.%
, Lee, C\BHBI Y.%
, Tippett, M\BPBI K.%
, Chavas, D\BPBI R.%
\BCBL {}\ \BBA {} Knutson, T\BPBI R.%
\end{APACrefauthors}%
\unskip\
\newblock
\APACrefYearMonthDay{2022}{{\APACmonth{04}}}{}.
\newblock
{\BBOQ}\APACrefatitle {Machine {Learning}–{Based} {Hurricane} {Wind} {Reconstruction}} {Machine {Learning}–{Based} {Hurricane} {Wind} {Reconstruction}}.{\BBCQ}
\newblock
\APACjournalVolNumPages{Weather and Forecasting}{}{}{}.
\newblock
\begin{APACrefURL} [{2024-10-01}]\url{https://journals.ametsoc.org/view/journals/wefo/37/4/WAF-D-21-0077.1.xml} \end{APACrefURL}
\newblock
\APACrefnote{Section: Weather and Forecasting}
\newblock
\begin{APACrefDOI} \doi{10.1175/WAF-D-21-0077.1} \end{APACrefDOI}
\PrintBackRefs{\CurrentBib}

\bibitem [\protect \citeauthoryear {%
Zorita%
\ \BBA {} Storch%
}{%
Zorita%
\ \BBA {} Storch%
}{%
{\protect \APACyear {1999}}%
}]{%
zorita1999}
\APACinsertmetastar {%
zorita1999}%
\begin{APACrefauthors}%
Zorita, E.%
\BCBT {}\ \BBA {} Storch, H\BPBI v.%
\end{APACrefauthors}%
\unskip\
\newblock
\APACrefYearMonthDay{1999}{{\APACmonth{08}}}{}.
\newblock
{\BBOQ}\APACrefatitle {The {Analog} {Method} as a {Simple} {Statistical} {Downscaling} {Technique}: {Comparison} with {More} {Complicated} {Methods}} {The {Analog} {Method} as a {Simple} {Statistical} {Downscaling} {Technique}: {Comparison} with {More} {Complicated} {Methods}}.{\BBCQ}
\newblock
\APACjournalVolNumPages{Journal of Climate}{}{}{}.
\newblock
\begin{APACrefURL} [{2025-03-06}]\url{https://journals.ametsoc.org/view/journals/clim/12/8/1520-0442_1999_012_2474_tamaas_2.0.co_2.xml} \end{APACrefURL}
\newblock
\APACrefnote{Section: Journal of Climate}
\PrintBackRefs{\CurrentBib}

\end{thebibliography}

\appendix
\section{}

\begin{figure}[!h]
\begin{center}
        \includegraphics[width=0.6\linewidth]{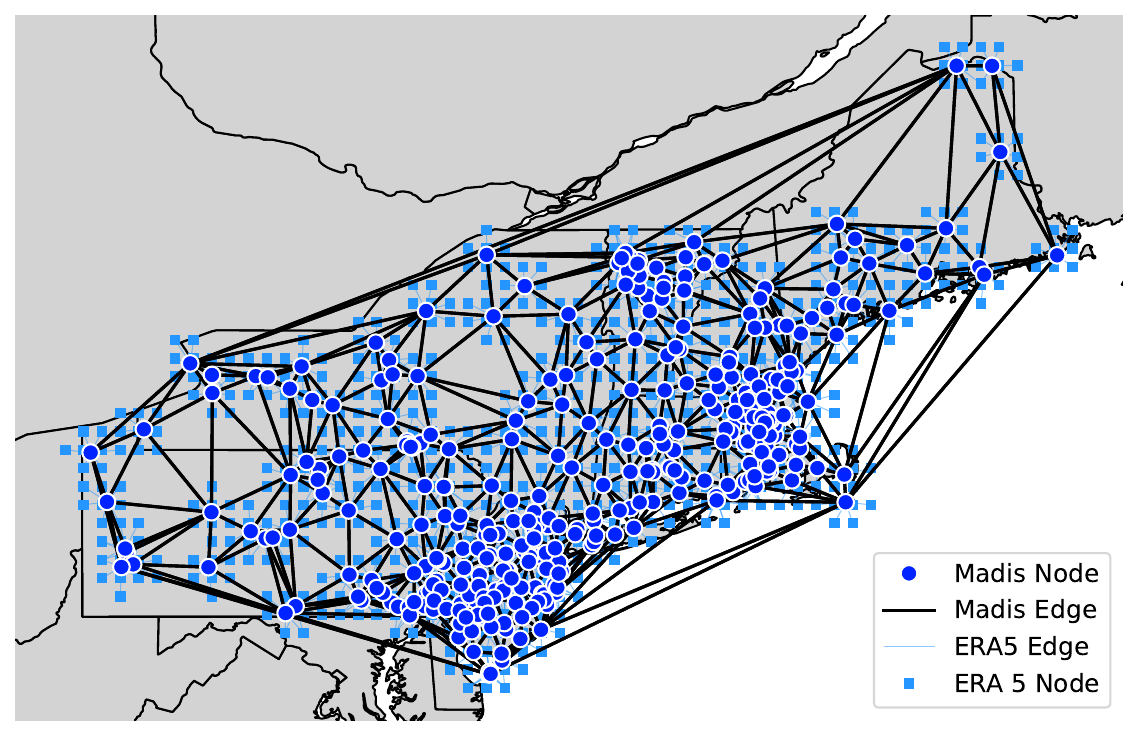}
    \end{center}
\caption{\textbf{Weather station and ERA5 graph for the US Northeast study region.} 
    MADIS stations are shown as dark blue circles; 
    the black edges connect each weather station in a Delaunay Network. 
    ERA5 nodes are shown as light blue squares, and each weather station is connected to its 8 nearest ERA5 neighbors.}
    \label{fig:methods:experiments}
\end{figure}

\begin{table}[!h]
    \caption{\textbf{Summary of our curated dataset.} This consists of three parts: ERA5, HRRR, and MADIS.}
    \label{tab:methods:dataset}
    \begin{center}
    \begin{tabular}{cccccc}
    \toprule
    Name & Type & Temporal Span & Spatial Span & Variables \\
    \midrule
    \multirow{4}{*}{ERA5} & \multirow{4}{*}{Gridded Mesh} & \multirow{4}{*}{2019--2023} & \multirow{4}{*}{Northeast US} & 10m $u$, 10m $v$, \\
     &  &  &  & 2m temperature, \\
     &  &  &  & 2m dewpoint temperature,\\
    \midrule
    \multirow{4}{*}{HRRR} & \multirow{4}{*}{Gridded Mesh} & \multirow{4}{*}{2019--2023} & \multirow{4}{*}{Northeast US} & 10m $u$, 10m $v$, \\
     &  &  &  & 2m temperature, \\
     &  &  &  & 2m dewpoint temperature,\\
    \midrule
    \multirow{5}{*}{MADIS} & \multirow{5}{*}{Off-Grid Station}  & \multirow{5}{*}{2019--2023} & \multirow{5}{*}{Northeast US} & 10m wind speed, \\
     &  &  &  & 10m wind direction,  \\
     &  &  &  & 2m temperature,  \\
     &  &  &  & 2m dewpoint temperature, \\
    \bottomrule
    \end{tabular}
    \end{center}
\end{table} 
\begin{table}[!h]
    \caption{\textbf{Test set Wind Vector Error [m/s] of experiments for each lead time.} Wind Vector Error is averaged over stations and time steps for the year 2023.}
    \label{tab:results:Wind}
\begin{center}
    \resizebox{\columnwidth}{!}{\begin{tabular}{cccccccccccc}
        \toprule
        Model & ExternalDataset &   1 &   2 &   4 &   8 &   12 &   18 &   24 &   36 &   48 &  Mean Error \\
        \midrule
        Interpolation &            ERA5 &        2.6206 &        2.6206 &        2.6206 &        2.6206 &         2.6206 &         2.6206 &         2.6206 &         2.6206 &         2.6206 &      2.6206 \\
        Interpolation &          HRRR-A &        2.4617 &        2.4617 &        2.4617 &        2.4617 &         2.4617 &         2.4617 &         2.4617 &         2.4617 &         2.4617 &      2.4617 \\
        Interpolation &          HRRR-F &        2.6235 &        2.6506 &        2.6659 &        2.6901 &         2.7177 &         2.7792 &         - &         - &         - &      2.6880 \\
        Persistence &               - &          0.3885 &        0.5165 &        0.6953 &        0.9373 &         1.0740 &         1.1415 &         1.1464 &         1.3364 &         1.2522 &      0.9431 \\
        \midrule
                MLP &               - &        0.3855 &        0.5012 &        0.6500 &        0.8090 &         0.8683 &         0.8943 &         0.9053 &         0.9261 &         0.9262 &      0.7629 \\
                MLP &            ERA5 &        0.4474 &        0.5366 &        0.6267 &        0.6930 &         0.6956 &         0.6844 &         0.6997 &         0.7136 &         0.7370 &      0.6482 \\
                MLP &          HRRR-F &        0.4252 &        0.5294 &        0.6228 &        0.6970 &         0.7247 &         0.7244 &         0.8399 &         0.9255 &         0.9265 &      0.7128 \\
                MLP &          HRRR-A &        0.4391 &        0.5229 &        0.6105 &        0.6721 &         0.6509 &         0.6667 &         0.6624 &         0.6729 &         0.6857 &      0.6204 \\
        \midrule
                GNN &               - &        0.4290 &        0.5235 &        0.6518 &        0.7806 &         0.8487 &         0.8821 &         0.9009 &         0.9258 &         0.9300 &      0.7636 \\
                GNN &            ERA5 &        0.4201 &        0.5146 &        0.6097 &        0.6991 &         0.7159 &         0.6941 &         0.7079 &         0.7248 &         0.7148 &      0.6446 \\
                GNN &          HRRR-F &        0.4171 &        0.5089 &        0.6123 &        0.6957 &         0.6991 &         0.7133 &         0.8482 &         0.9119 &         0.9288 &      0.7039 \\
                GNN &          HRRR-A &        0.4248 &        0.5015 &        0.5969 &        0.6721 &         0.7030 &         0.6661 &         0.6700 &         0.6849 &         0.6886 &      0.6231 \\
        \midrule
        Transformer &               - &        0.3782 &        0.4838 &        0.6040 &        0.7216 &         0.7791 &         0.8515 &         0.8867 &         0.9113 &         0.9123 &      0.7254 \\
        Transformer &            ERA5 &        0.3736 &        0.4444 &        0.5081 &        0.5185 &         0.5284 &         0.5260 &         0.5354 &         0.5388 &         0.5351 &      0.5009 \\
        Transformer &          HRRR-F &        0.3791 &        0.4458 &        0.4943 &        0.5335 &         0.5464 &         0.5617 &         0.7019 &         0.8713 &         0.9130 &      0.6052 \\
        Transformer &          HRRR-A &        0.3759 &        0.4364 &        0.4819 &        0.4948 &         0.5041 &         0.5086 &         0.5055 &         0.5160 &         0.5224 &      0.4828 \\
        \bottomrule
    \end{tabular}
    }
    \end{center}
\end{table} \begin{table}[!h]
    \caption{\textbf{Test set Temperature RMSE [$^{\circ}$C] of experiments for each lead time.} Temperature RMSE is averaged over stations and time steps for the year 2023.}
    \label{tab:results:Temp}
\begin{center}
    \resizebox{\columnwidth}{!}{\begin{tabular}{cccccccccccc}
        \toprule
        Model & ExternalDataset &   1 &   2 &   4 &   8 &   12 &   18 &   24 &   36 &   48 &  Mean Error \\
        \midrule
        Interpolation &            ERA5 &       2.0285 &       2.0285 &       2.0285 &       2.0285 &        2.0285 &        2.0285 &        2.0285 &        2.0285 &        2.0285 &      2.0285 \\
        Interpolation &          HRRR-A &       1.6824 &       1.6824 &       1.6824 &       1.6824 &        1.6824 &        1.6824 &        1.6824 &        1.6824 &        1.6824 &      1.6824 \\
        Interpolation &          HRRR-F &       1.7706 &       1.8379 &       1.9356 &       2.0467 &        2.1033 &        2.1501 &           - &           - &           - &            1.9740 \\
        Persistence &               - &         1.1285 &       2.0531 &       3.5960 &       5.6626 &        6.4681 &        5.6292 &        4.3057 &        7.3789 &        5.3896 &      4.6235 \\
        \midrule
                MLP &               - &        0.8586 &        1.3821 &        2.1692 &        3.0583 &         3.4384 &         3.7497 &         3.9832 &         4.6830 &         4.8350 &      3.1286 \\
                MLP &            ERA5 &        1.6848 &        1.7721 &        2.1847 &        2.6728 &         2.6778 &         2.5402 &         2.5701 &         2.5190 &         2.2849 &      2.3229 \\
                MLP &          HRRR-F &        1.5682 &        2.2566 &        2.1774 &        2.8183 &         2.7663 &         2.9323 &         3.0804 &         3.5388 &         4.2048 &      2.8159 \\
                MLP &          HRRR-A &        1.5194 &        1.9444 &        2.0540 &        2.4136 &         2.6902 &         2.0723 &         2.1444 &         2.0465 &         2.3971 &      2.1424 \\
        \midrule
                GNN &               - &        1.6763 &        2.0039 &        2.4243 &        2.8027 &         3.3179 &         3.3818 &         3.9041 &         4.4925 &         5.0684 &      3.2302 \\
                GNN &            ERA5 &        1.7567 &        2.0910 &        2.2112 &        2.7139 &         2.8009 &         2.3816 &         2.5855 &         2.8181 &         2.2044 &      2.3959 \\
                GNN &          HRRR-F &        1.7773 &        1.8839 &        2.8742 &        2.5763 &         2.8597 &         2.7491 &         3.0851 &         3.6259 &         4.1813 &      2.8459 \\
                GNN &          HRRR-A &        1.7077 &        1.9705 &        2.0594 &        2.4264 &         2.7071 &         2.6259 &         3.0305 &         2.8871 &         2.4290 &      2.4271 \\
        \midrule
        Transformer &               - &        0.8453 &        1.3104 &        1.9738 &        2.6415 &         2.9751 &         3.3225 &         3.6866 &         4.3659 &         4.5679 &      2.8543 \\
        Transformer &            ERA5 &        0.7850 &        1.1293 &        1.4047 &        1.5725 &         1.5897 &         1.5775 &         1.6328 &         1.6226 &         1.6761 &      1.4434 \\
        Transformer &          HRRR-F &        0.7890 &        1.1146 &        1.4106 &        1.6805 &         1.7700 &         1.8012 &         2.4541 &         3.3228 &         4.1414 &      2.0538 \\
        Transformer &          HRRR-A &        0.7709 &        0.9916 &        1.2261 &        1.3163 &         1.3852 &         1.3612 &         1.3775 &         1.4410 &         1.4685 &      1.2598 \\
        \bottomrule
    \end{tabular}
    }
    \end{center}
\end{table} \begin{table}[!h]
    \caption{\textbf{Test set Dewpoint RMSE [$^{\circ}$C] of experiments for each lead time.} Dewpoint RMSE is averaged over stations and time steps for the year 2023.}
    \label{tab:results:Dew}
\begin{center}
    \resizebox{\columnwidth}{!}{\begin{tabular}{cccccccccccc}
        \toprule
        Model & ExternalDataset &  1 &  2 &  4 &  8 &  12 &  18 & 24 &  36 &  48 &  Mean Error \\
        \midrule
        Interpolation &            ERA5 &       3.5549 &       3.5549 &      3.5549 &      3.5549 &       3.5549 &       3.5549 &       3.5549 &       3.5549 &       3.5549 &     3.5549 \\
        Interpolation &          HRRR-A &       3.4996 &       3.4996 &      3.4996 &      3.4996 &       3.4996 &       3.4996 &       3.4996 &       3.4996 &       3.4996 &     3.4996 \\
        Interpolation &          HRRR-F &       3.5716 &       3.5926 &      3.6243 &      3.6774 &       3.7137 &       3.7326 &            - &            - &            - &     3.6520 \\
        Persistence &               - &         0.7834 &       1.2740 &      2.0021 &      3.0099 &       3.6988 &       4.3591 &       4.8231 &       5.8531 &       6.1053 &     3.5454 \\
        \midrule
                MLP &               - &        0.7919 &        1.2483 &        1.8765 &        2.7753 &         3.3775 &         3.9771 &         4.3993 &         5.0861 &         5.3832 &      3.2128 \\
                MLP &            ERA5 &        1.4267 &        1.6521 &        1.9811 &        2.5329 &         2.9191 &         2.7480 &         2.8060 &         2.7717 &         2.5846 &      2.3802 \\
                MLP &          HRRR-F &        1.4384 &        2.0339 &        2.0095 &        2.5592 &         2.7156 &         2.8836 &         2.9783 &         3.6511 &         4.5118 &      2.7535 \\
                MLP &          HRRR-A &        1.3789 &        1.6538 &        2.0259 &        2.4997 &         2.7184 &         2.4792 &         2.4657 &         2.4913 &         2.7995 &      2.2792 \\
        \midrule
                GNN &               - &        1.4278 &        1.8289 &        2.0423 &        2.5904 &         2.9936 &         3.4755 &         3.9815 &         4.8123 &         5.3336 &      3.1651 \\
                GNN &            ERA5 &        1.4616 &        1.7783 &        2.2188 &        2.4419 &         2.7946 &         2.6365 &         2.8001 &         3.0051 &         2.4818 &      2.4021 \\
                GNN &          HRRR-F &        1.4717 &        1.8274 &        2.6824 &        2.3998 &         3.0036 &         2.7710 &         3.0227 &         3.7669 &         4.4572 &      2.8225 \\
                GNN &          HRRR-A &        1.7071 &        1.6215 &        1.9472 &        2.4414 &         2.7491 &         2.6551 &         3.0125 &         2.9933 &         2.6584 &      2.4206 \\
        \midrule
        Transformer &               - &        0.7721 &        1.1771 &        1.7359 &        2.3718 &         2.6763 &         3.4317 &         3.9042 &         4.7105 &         5.0570 &      2.8707 \\
        Transformer &            ERA5 &        0.7820 &        1.1003 &        1.3715 &        1.6132 &         1.6945 &         1.7553 &         1.8099 &         1.8821 &         1.9115 &      1.5467 \\
        Transformer &          HRRR-F &        0.7566 &        1.1658 &        1.3903 &        1.6845 &         1.8506 &         1.9727 &         2.4297 &         3.5142 &         4.4056 &      2.1300 \\
        Transformer &          HRRR-A &        0.7634 &        1.0256 &        1.3041 &        1.4809 &         1.6162 &         1.6522 &         1.7129 &         1.7841 &         1.8297 &      1.4632 \\
        \bottomrule
    \end{tabular}
    }
    \end{center}
\end{table} 
\begin{figure}[htbp]
    \begin{center}
        \includegraphics[width=1\linewidth]{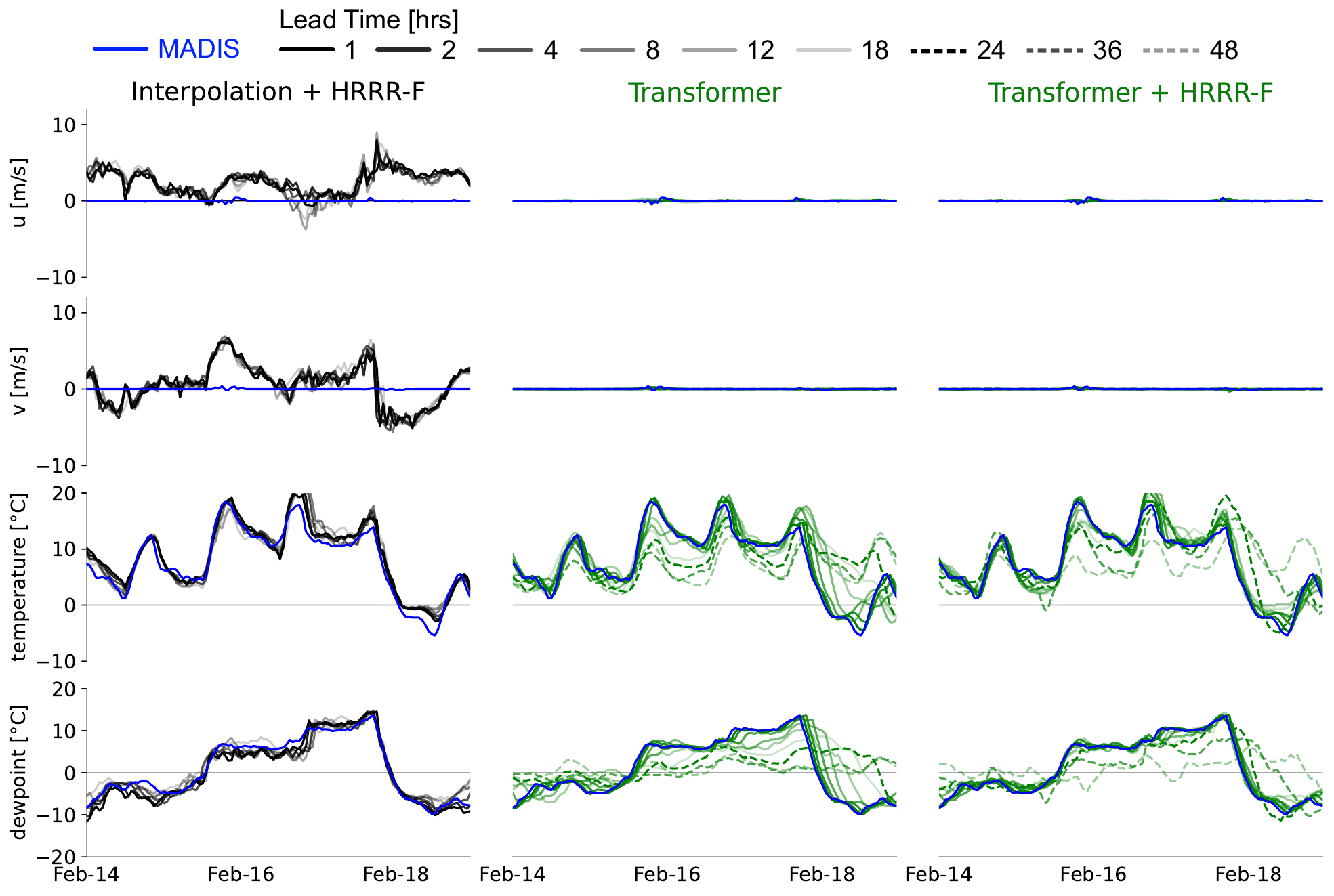}
    \end{center}
    \caption{
        \textbf{Example time series from one weather station where transformer + HRRR-F shows the smallest error.}
        From top to bottom: time series of 10m $u$ and $v$ components, 2m temperature, and 2m dewpoint are presented.
        From left to right: HRRR forecast interpolation, transformer with MADIS only, and transformer with additional HRRR forecasts are demonstrated.
        Each panel has the MADIS ground truth in blue, and the predictions at increasing lead times displayed with decreasing saturation.
        For this station and time snippet the interpolated HRRR-F appears rather inaccurate for wind vector estimation.
        The wind vector value is very close to zero suggesting wind blocking effect from local topography, which is extremely hard to be captured by large-scale numerical model like HRRR. 
        By ingesting history of local measurements, the transformer is able to easily capture this local heterogeneity.
    }\label{fig:results:time_series_T_HRRR_F_good}
    \vspace{-.7cm}
\end{figure} \begin{figure}[!h]
    \begin{center}
        \includegraphics[width=1\linewidth]{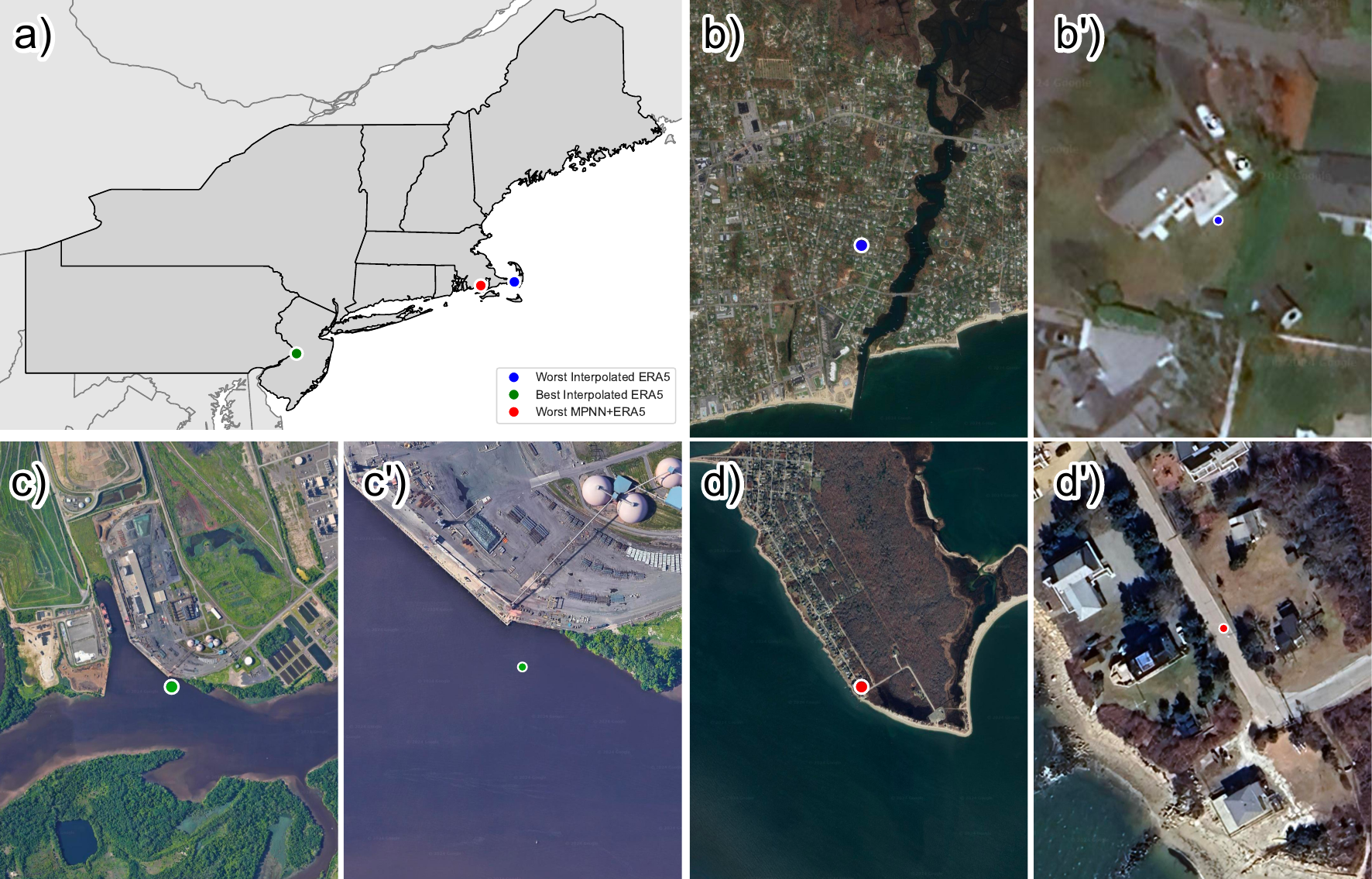}
    \end{center}
    \caption{
        \textbf{Examples of environments for two selected weather stations.}
        (a) Overview of the two stations. (b) Google Satellite image of the two selected weather stations.
        The two selected weather stations are: (c) the station with the best overall Transformer + HRRR-F loss, (d) and the station with the worst overall Transformer + HRRR-F loss.
        (c) and (d) show a zoomed-out view of each area; (c') and (d') are zoomed in.
        (c) shows good examples of weather stations being surrounded by trees and buildings, where wind is affected by these local surface characteristics.
    }\label{fig:appendix:stations_comparison}
\end{figure} 
\section*{Model Details}

\subsection{Model Complexity}
In terms of time complexity, GNN scales as $O(kn)$ and transformer scales as $O(n^2)$, 
where $n$ is the number of nodes (or tokens) and $k$ is the number of neighbors connected to each node.
In terms of space complexity, GNN and transformer also scale as $O(kn)$ and $O(n^2)$, separately. 
As a result, GNN will be much more efficient when the underlying graph structure is sparse.

\subsection{Hyperparameters}

All models were trained with the Adam optimizer~\cite{kingma2017} with a learning rate of $10^{-4}$, a weight decay of $10^{-4}$, and a batch size of 128.
All models were trained for 200 epochs to ensure model convergence, and the best model was selected based on the validation set performance.

\begin{table}[!h]
    \caption{
        \textbf{Ablation study on model structure.}
        Comparison of errors for the three variables when varying one of the parameters (for a single lead time - 12 hours): network connection between MADIS weather stations, message passing steps (only for GNN), and number of connected ERA5 nodes.
        In the network structure column, \textit{Fully Connected} means that every node is connected to every other node, while \textit{Delaunay} refers to the Delaunay triangulation of the weather stations, an optimized way of connecting nodes based on their spatial location.
        In the message passing steps column, the number indicates how many times messages are passed between local nodes during the GNN training process.
        \textit{Diameter} refers to the number of hops (here message passing steps) needed to connect any two pairs of nodes in the graph, ensuring the all information can propagate.
        In the connected ERA5 nodes column, the number indicates how many ERA5 nodes are connected to the GNN.
        \textit{Stacked} means that the nodes closest to each weather station are stacked together before being tokenized.
        The first section shows the results reported in the main text, while the second section shows additional results with different configurations.
        \textbf{Number of message passing steps:} The number of message passing steps does not seem to significantly improve the performance of the model. Only a small improvement is seen for the wind and dewpoint errors compared to the results in the main text, and only for wind when limiting the number of connected ERA5 nodes to 1.
        \textbf{Network connection:} The fully connected GNN slightly improves the performance for all variables, particularly when limiting the number of connected ERA5 nodes to 1, and now achieves a better performance than the MLP model.
        \textbf{Neighborhood:} Limiting the number of connected ERA5 nodes to 1 also improves the performance of almost all models comparing to their counterparts with 8 ERA5 nodes, which also allows almost all of them to beat the performance of the MLP model.
        Overall, it seems that the model gets confused by the increased complexity of the input data when more ERA5 nodes are connected, which is also confirmed by the experiment on the Transformer model.
        When the Transformer is fed a stack of gridded weather data from multiple ERA5 nodes, it struggles to effectively utilize the additional information, leading to degraded performance.
        Overall though, it still outperforms the GNN and MLP.
        }
    \label{tab:ablation:structure}
    \begin{center}
    \resizebox{\columnwidth}{!}{\begin{tabular}{llccccc}
\toprule
Model & \shortstack[l]{Weather Stations\\Network Structure} & \shortstack[c]{Message Passing\\Steps} & \shortstack[c]{Connected\\ERA5 Nodes} & Wind [m/s] & \shortstack[c]{Error\\Temperature [$^{\circ}$C]} & Dewpoint [$^{\circ}$C] \\
\midrule
MLP & Single Node & - & 1 & 0.6956 & 2.6778 & 2.9191 \\
Transformer & Fully Connected & - & 1 & 0.5284 & 1.5897 & 1.6945 \\
GNN & Delaunay & 4 & 8 & 0.7159 & 2.8009 & 2.7946 \\\hline
GNN & Delaunay & 14 (Diameter) & 8 & 0.7039 & 2.8130 & 2.7580 \\
GNN & Fully Connected & 1 & 8 & 0.7021 & 2.2943 & 2.4937 \\
GNN & Delaunay & 4 & 1 & 0.7149 & 2.6725 & 2.6376 \\
GNN & Delaunay & 14 (Diameter) & 1 & 0.6982 & 2.9867 & 2.8098 \\
GNN & Fully Connected & 1 & 1 & 0.6681 & 2.2819 & 2.4544 \\
Transformer & Fully Connected & - & 8 (Stacked) & 0.5856 & 1.6555 & 1.9060 \\
\bottomrule
\end{tabular}
     }
    \end{center}
\end{table}

\end{document}